\documentclass[10pt,journal,compsoc]{IEEEtran}
\ifCLASSOPTIONcompsoc
  \usepackage[nocompress]{cite}
\else
  \usepackage{cite}
\fi
\ifCLASSINFOpdf
\else
\fi
\usepackage{amsmath,amsfonts}
\usepackage{algorithmic}
\usepackage{algorithm}
\usepackage{array}
\usepackage[caption=false,font=normalsize,labelfont=sf,textfont=sf]{subfig}
\usepackage{textcomp}
\usepackage{stfloats}
\usepackage{url}
\usepackage{verbatim}
\usepackage{graphicx}
\usepackage{cite}

\usepackage{color,array}

\usepackage{graphicx}
\usepackage{multirow}
\usepackage[flushleft]{threeparttable}
\usepackage{float}

\usepackage{subfig}

\usepackage{makecell}
\usepackage{color,soul}
\usepackage{xcolor}
\definecolor{cerulean}{rgb}{0.0,0.48,0.65}
\definecolor{green}{rgb}{0.01, 0.75, 0.24}
\definecolor{Black}{RGB}{0.0, 0.0, 0.0}

\newcommand{\brown}[1]{\textcolor{Black}{#1}}

\newcommand{\shadow}[1]{}

\def\br{\brown}

\begin{document}
\title{Graph Anomaly Detection in Time Series: \\ A Survey}

\author{Thi Kieu Khanh Ho, Ali Karami, Narges Armanfard% <-this % stops a space
\IEEEcompsocitemizethanks{\IEEEcompsocthanksitem TKK. Ho, A. Karami, and N. Armanfard are with McGill University and Mila-Quebec AI Institute, Montreal, Canada.\protect\\

\IEEEcompsocthanksitem E-mails: thi.k.ho@mail.mcgill.ca, ali.karami@mail.mcgill.ca, narges.armanfard@mcgill.ca.}}% <-this % stops an unwanted space
% \thanks{Manuscript received April 19, 2005; revised August 26, 2015.}}


\markboth{Journal of \LaTeX\ Class Files,~Vol.~14, No.~8, August~2015}%
{Shell \MakeLowercase{\textit{et al.}}: Bare Demo of IEEEtran.cls for Computer Society Journals}



\IEEEtitleabstractindextext{%
\begin{abstract}
 With the recent advances in technology, a wide range of systems continue to collect a large amount of data over time and thus generate time series. Time-Series Anomaly Detection (TSAD) is an important task in various time-series applications such as e-commerce, cybersecurity, vehicle maintenance, and healthcare monitoring. However, this task is very challenging as it requires considering both the \textit{intra-variable dependency} (relationships within a variable over time) and the \textit{inter-variable dependency} (relationships between multiple variables) existing in time-series data. Recent graph-based approaches have made impressive progress in tackling the challenges of this field. In this survey, we conduct a comprehensive and up-to-date review of TSAD using graphs, referred to as G-TSAD. First, we explore the significant potential of graph representation for time-series data and and its contributions to facilitating anomaly detection. Then, we review state-of-the-art graph anomaly detection techniques, mostly leveraging deep learning architectures, in the context of time series. For each method, we discuss its strengths, limitations, and the specific applications where it excels. Finally, we address both the technical and application challenges currently facing the field, and suggest potential future directions for advancing research and improving practical outcomes.
\end{abstract}

}

\maketitle

\IEEEdisplaynontitleabstractindextext

% \fi
\IEEEpeerreviewmaketitle

\IEEEraisesectionheading{\section{Introduction}\label{sec:introduction}}


\IEEEPARstart{A} time series is defined as an ordered sequence of  values that represent the evolution of one variable, aka \textit{univariate}, or multiple variables, aka \textit{multivariate}, over time \cite{schmidl2022anomaly}. Time-series data is also referred to as time-stamped data, i.e., each data point or each observation corresponds to a specific time index. As time is a constituent of everything that is observable, time-series data can be found everywhere on the Earth, such as in nature (e.g., the wind speed, the temperature), in marketing and industrial activities (e.g., the stock price), in medicine (e.g., the heart and brain activity), and in surveillance and security (e.g., video streams, network traffic logs).  Therefore, time-series data and its analysis have attracted significant interest in the past decade.

Time-Series Anomaly Detection (TSAD), the process of detecting unusual patterns that do not conform to the expected behavior, has been widely studied \cite{blazquez2021review}. The unusual patterns can be found in many real-world applications such as ecosystem disturbances in earth sciences, structural defects in jet turbine engineering, suspicious activities in surveillance videos, heart failure in cardiology, or seizures in brain activity. Most existing TSAD algorithms are designed to analyze time-series data as a univariate modality \cite{zong2018deep,ren2019time,kieu2019outlier,lai2021revisiting,rebjock2021online,wu2021current,kim2022towards,deng2022cadet,hojjati2022selfsig,morais2019learning,feng2021mist,georgescu2021anomaly,huang2022self,ristea2022self,wu2022self} or as multiple individual modalities (variables) \cite{zhang2019deep,zhou2019beatgan,audibert2020usad,zhang2021unsupervised,abdulaal2021practical,chen2021daemon,zhang2021unsupervised,tuli2022tranad}. In other words, these algorithms only consider the intra-variable information.

In many practical systems, a more proper and reliable system modeling can be obtained if the  data are considered as multiple related modalities (variables) that are influenced by each other. Specifically, a comprehensive anomaly detection (AD) algorithm carefully shall model the variables' interaction over time, i.e., the inter-variable dependencies, as well as the intra-variable dependencies. Explicit consideration of the inter-variable dependencies allows capturing a system's spatial dependencies across variables. Examples of systems, which exhibit spatial dependencies, are the brain where the recording sensors are placed across the brain, and each sensor is considered as a variable; in a single time-series signal where every sample (feature) or time-interval (window) is considered as a variable; and in a video-based system where a variable can be assigned to every frame. 

%In the current literature, depending on the application in hand, the researchers refer the inter-variable dependency as the spatial dependency \cite{deng2022graph,chen2022deep}. On the other hand, existing studies typically refer to the intra-variable dependency as the temporal dependency \cite{dai2022graph,chen2021learning}.

While many TSAD algorithms have addressed intra-variable dependencies, capturing inter-variable dependencies remains under-explored in existing studies. In recent years, graph-based methods have gained popularity in TSAD due to their ability to represent complex data structures inherent in time-series data. Unlike traditional tabular representations, graphs have demonstrated superior performance in AD by modeling inter-variable dependencies, where each variable is represented as a graph node and the interactions between variables are depicted as graph edges \cite{liu2022graph,wu2021self}. This property is valuable for TSAD. For instance, if one variable experiences a change, related variables are likely to exhibit similar changes as well. This enables the identification of anomalies that may not be evident through conventional TSAD methods.

This survey explores TSAD using graph-based approaches. It categorizes existing research based on the types of constructed graphs, the nature of graph-based anomalies in time-series data, the detection methods employed, and the application domains.

\textbf{Paper Organization.} Section \ref{sec:surveys} presents an overview of existing surveys in the literature and our contributions. Section \ref{sec:timeseries} outlines the challenges in time-series data. Sections \ref{sec:graphs} and \ref{sec:typesAno} provide foundational insights into graphs, common types of anomalies within graphs, and how the graph-based methods can address the time-series challenges. Section \ref{sec:methods} introduces our unified taxonomy for TSAD using graphs. Within each category, we discuss the strength of each method in addressing the challenges outlined in Section \ref{sec:timeseries}, as well as its applications and technical drawbacks and how the subsequent methods mitigate these shortcomings. Section \ref{sec:datasets} summarizes widely used datasets and their real-world applications. Section \ref{sec:evaluation_metric} discusses employed evaluation metrics in the field. Section \ref{sec:discussion} discusses the current research challenges and promising directions for future research. Finally, we conclude our paper in Section \ref{sec:conclusion}.

\section{Overview of Related Surveys} \label{sec:surveys}

Although graphs have proven effective and yielded state-of-the-art results in TSAD, as summarized in Table \ref{tab:surveys}, {\textit{none of existing surveys have focused on the emergence and applications of graphs for detecting anomalies in time-series data}, referred to as G-TSAD. For instance, Graph-based Approaches (GA) surveys \cite{wu2020comprehensive,wu2021self,liu2022graph,xie2022self,zhou2022graph,li2022graph,liu2022survey,zhang2022few,gao2023survey,khoshraftar2024survey,jin2024survey} cover a broad range of graph structure learning techniques for various tasks such as classification, clustering, and prediction, without focusing specifically on AD.  On the other hand, Graph Anomaly Detection (GAD) surveys \cite{ma2021comprehensive,kim2022graph,pazho2023survey,ren2023graph,bilot2023graph,lamichhane2024anomaly,ekle2024anomaly} provides a comprehensive review of graph-based methods for detecting various types of abnormal graph objects. However, both GA and GAD surveys solely review the graph-based methods applicable to static data (e.g., static social networks or images) and do not provide comprehensive reviews on the methods that are able to capture the intra-variable dependencies existing in time-series data. Meanwhile, TSAD surveys \cite{cook2019anomaly,choi2021deep,blazquez2021review,li2023deep,zamanzadeh2024deep} focus on methodologies for detecting anomalies within time-series signals, disregarding other data types of the time series such as dynamic social networks and videos. Additionally,} existing TSAD surveys show the ability of non-graph deep learning methods to detect anomalies only based on the intra-variable dependencies, ignoring the inter-variable dependencies; this is while both intra- and inter-variable dependencies are crucial for TSAD.

\begin{table}[!t]
\caption{Comparison of the existing surveys with ours. $\star$ indicates that the topic is reviewed in the corresponding survey study, -- indicates otherwise.}
\fontsize{9}{8.4}\selectfont
\centering
\begin{tabular}{c c | c | c | c | c} 
 \hline
 \textbf{Survey} & Year & GA & GAD & TSAD & G-TSAD \\ 
 \hline
  \cite{wu2020comprehensive} & 2020 & $\star$ & -- & -- & -- \\ 
  \cite{wu2021self} & 2021 & $\star$ & -- & -- & -- \\
  \cite{xia2021graph} & 2021 & $\star$ & -- & -- & -- \\
  \cite{liu2022graph} & 2022 & $\star$ & -- & -- & -- \\ 
  \cite{xie2022self} & 2022 & $\star$ & -- & -- & -- \\
  \cite{zhou2022graph} & 2022 & $\star$ & -- & -- & -- \\
  \cite{li2022graph} & 2022 & $\star$ & -- & -- & -- \\
  \cite{liu2022survey} & 2022 & $\star$ & -- & -- & -- \\
  \cite{zhang2022few} & 2022 & $\star$ & -- & -- & -- \\
  \cite{gao2023survey} & 2023 & $\star$ & -- & -- & -- \\
  \cite{khoshraftar2024survey} & 2024 & $\star$ & -- & -- & -- \\
  \cite{jin2024survey} & 2024 & $\star$ & -- & -- & -- \\
  \cite{ma2021comprehensive} & 2021 & -- & $\star$ & -- & -- \\
  \cite{kim2022graph} & 2022 & -- & $\star$ & -- & -- \\
  \cite{pazho2023survey} & 2023 & -- & $\star$ & -- & -- \\
  \cite{ren2023graph} & 2023 & -- & $\star$ & -- & -- \\
  \cite{bilot2023graph} & 2023 & -- & $\star$ & -- & -- \\
  \cite{lamichhane2024anomaly} & 2024 & -- & $\star$ & -- & -- \\
  \cite{ekle2024anomaly} & 2024 & -- & $\star$ & -- & -- \\
  \cite{cook2019anomaly} & 2019 & -- & -- & $\star$ & -- \\
  \cite{choi2021deep} & 2021  & -- & -- & $\star$ & -- \\
  \cite{blazquez2021review} & 2021 & -- & -- & $\star$ & -- \\
  \cite{li2023deep} & 2023  & -- & -- & $\star$ & -- \\
  \cite{zamanzadeh2024deep} & 2024  & -- & -- & $\star$ & -- \\
  \hline
  \textbf{Ours} & 2024 & -- & -- & -- & $\star$ \\ 
 \hline
\end{tabular}
\label{tab:surveys}
\end{table}

The contributions of this survey are as follows:
\begin{itemize}
    \item \textbf{The first survey in G-TSAD}. To the best of our knowledge, this is the first survey that reviews the state-of-the-art graph-based techniques for TSAD. Our primary goal of this survey is to showcase how anomaly detectors, which leverage graphs, detect anomalies in time-series data. Until now, all the relevant surveys focus on either GA, TSAD, or GAD. There is no dedicated and comprehensive survey on G-TSAD. Our work bridges this gap, and we expect that such structured and comprehensive survey helps to push forward the research in this active area.
    \item \textbf{A new perspective}. We introduce new concepts in the field of G-TSAD, such as the intra- and inter-variable dependencies, and constructed graphs for time-series data. Such concepts would pave the road for future research to provide well-structured and easy-to-understand G-TSAD algorithms.
    \item \textbf{A detailed overview of G-TSAD}. We discuss the key challenges in TSAD, the core motivations of employing graph representation for time-series data and its role in facilitating anomaly detection, and the various types of anomalies in graphs. 
    \item \textbf{A systematic and comprehensive review}. We conduct a comprehensive and up-to-date review of the state-of-the-art on G-TSAD methods, primarily leveraging deep learning architectures, and categorize them into four groups to improve their clarity and accessibility. These categories include Autoencoder (AE)-based methods, Generative Adversarial Network (GAN)-based methods, predictive-based methods, and self-supervised methods. For each category, we describe technical details, highlight their advantages and disadvantages, and compare them.
    \item \textbf{Outlook on future directions}. We point out both technical and application limitations of the current studies and suggest promising directions for future works on G-TSAD.
\end{itemize}

\section{Time-series Challenges} \label{sec:timeseries}

In general, a $K$-variable time series dataset can be denoted by $X = (\mathbf{x}^{(1)}, \mathbf{x}^{(2)}, \ldots, \mathbf{x}^{(K)})$, where each $\mathbf{x}^{(i)} = (x^{(i)}_1,x^{(i)}_2,\ldots, x^{(i)}_N)$, $X \in \mathbb{R}^{K \times N \times m}$, $\mathbf{x}^{(i)} \in \mathbb{R}^{N \times m}$, $K$ is the number of variables, $N$ is the number of observations, and $m$ is the feature length, representing the number of features in each observation. Each $x^{(i)}_j, j \in \{1,\ldots,N\}$ and $i \in \{1,\ldots,K\}$, is an $m$ dimensional vector corresponding to a recording at the $j$th time interval and the $i$th variable. For example, in Fig. \ref{fig:gtsad} (Left), the raw input data of three intervals of a dataset $X$ consisting of five variables (sensors) is shown. The raw interval at the $i$th variable is encoded by $m = 3$ features, as depicted on the top of each sensor in Block 2 of Fig. \ref{fig:gtsad}. Hence, $X$ can be represented as $X = (\mathbf{x}^{(1)}, \mathbf{x}^{(2)}, \ldots, \mathbf{x}^{(5)})$, where $X \in \mathbb{R}^{5\times 3 \times m}$, $\mathbf{x}^{(i)}=(x^{(i)}_1,\ldots, x^{(i)}_{j-1}, x^{(i)}_{j}, x^{(i)}_{j+1},\ldots, x^{(i)}_{N})$, $ \mathbf{x}^{(i)} \in \mathbb{R}^{3 \times m}$. Note that one may directly use the raw values as features; e.g., if an observation consists of 100 samples, then $m=100$. Similarly, a video sequence can be considered as either univariate or multivariate data. For univariate data, one may consider every frame or a subset of frames as one observation of that single variable, and use the raw pixel values of the frame(s) or their corresponding features. In the multivariate case, each pixel can be considered as a variable. In this context, an observation of a variable can be obtained from a single frame or a subset of frames.
% \s{(\b{or a constructed} feature) or a time interval (window) in the signals, a frame or sub-frames in a video sequence, or a snapshot in a dynamic social network.}

Working with time-series data requires careful consideration of several factors, which are discussed below in detail.

\textbf{Intra-variable dependency}. Observations available in one variable depend on each other,  i.e., there exist dependencies of the $j$th observation and a lagged version of itself in the $i$th variable. Positive dependencies indicate that an increase (or decrease) in the $j$th observation is likely caused by an increase (or decrease) in the previous observations and to be followed by an increase (or decrease) in the subsequent observations of the $i$th variable. In contrast, negative dependencies suggest an inverse relationship. However, in practice, one of the main obstacles is dealing with the non-linear complex dependencies present within a variable. In fact, the mutual relationship between observations is not straightforward, and the effect of the future/past observations on the current one may vary over time. This makes it difficult to determine the appropriate lags for consideration, as different lags may have different levels of significance. 

\textbf{Inter-variable dependency}. It is important to capture the inter-variable dependency as it can provide insights into the underlying relationship patterns that exist between different variables; a useful property for TSAD. For example, if two variables are correlated, a change in one variable can be used to predict changes in another variable. Another example is the case if individual variables have fairly normal behavior, but when all (or a subset) of variables and their interactions are considered together, an anomaly can be caught. Hence, employing graphs that can capture complex relationships between variables would be a great tool for TSAD.

\textbf{Non-stationarity}. It is important to consider trends, seasonality, and unpredictability. Although the trend is known as an average tendency of the series, the trend itself may vary over different time intervals. Seasonality presents the variations of repeating short-term cycles in the series, which causes a change of variance over time. Unpredictability occurs randomly that are neither systematic nor predictable. Such properties influence the variables statistical properties such as mean, variance, etc. Thus, time-series data is said to be non-stationary, which may easily mislead an AD method as anomalies at certain observations may not be true anomalies \cite{kim2022towards}. Methods that can adapt to the changes in the data structures require a large number of training data. 

\textbf{Dimensionality}. The current advances in technology allow us to record a large amount of time-series data to capture the intra-variable and inter-variable dependencies. Existing of such rich datasets allows us to design time-series analyses that are consistent and reliable across various datasets. Looking at the raw data, every data sample in every variable can be considered as one dimension; hence we are dealing with the curse of dimensionality challenge. Therefore, there is a need to develop algorithms that can handle such complicated and high-dimensional data.

\textbf{Noise}. It is important to make a distinction between noise and anomalies. Noise is the random, unwanted variation that degrades the data quality. In contrast, an anomaly is an unusual entity with respect to a set of pre-established normal observations. Therefore, AD algorithms explicitly designed to distinguish noise from anomalies are more robust when dealing with contaminated data; however, various types of noises may happen depending on the operational environment, recording sensor types, etc., that a researcher may not be aware at the problem onset. Hence, it is critical to understand the nature of noise in every application and apply appropriate noise modeling/reduction techniques to not mix them with the anomaly concept. 

In the subsequent sections, we will delve into graph-based techniques and their effectiveness in tackling individual challenges or combinations thereof present in time-series data, hence enhancing our understanding of their applicability and utility in real-world scenarios.

\section{Why Graphs?} \label{sec:graphs}

\subsection{Graph-Based Solutions for TSAD Challenges}

%While some recent studies have shown the inefficiency of the non-graph deep learning methods for TSAD \cite{kim2022towards,audibert2022deep,garg2021evaluation}, due to their utilization of unreliable benchmark datasets and biased evaluation protocols,

\subsubsection{Capturing Intra- and Inter-variable Dependencies} \label{subsec:constructed_graphs}

Many TSAD studies have shown superior performance when employing graphs. This is because many real-world systems can be represented as graphs, and graphs can inherently capture \textbf{intra- and inter-variable dependencies}. We define a graph set representing a time-series data as:
\begin{equation}
   \mathbb{G} = \Bigl\{\mathcal{G}_{j}, \text{Sim}\{\mathcal{G}_{j}, \mathcal{G}_{j'}\}\Bigl\}^{j, j' \in \{1, \ldots N\}}_{j \neq j'},
\end{equation}
where $\mathcal{G}_{j} = \{\mathbf{M}_{j}, \mathbf{A}_{j}\}$ is a graph representing the $j$th observation, $\mathbf{M}_{j}$ is the $K \times m$ node-feature matrix, $\mathbf{A}_{j}$ is the $K \times K \times m'$ edge-feature matrix representing the relationships across nodes (variables) in the $j$th observation, $m$ and $m'$ respectively denotes the number of features representing the nodes and their relations. $\text{Sim}\{\cdot,\cdot\}$ defines the relation between two graphs. The $\text{Sim}\{\cdot,\cdot\}$ goal is to capture the intra-variable dependencies between two observations. This can be defined as a function of node-features and edge-features, i.e., $\text{Sim}\{\mathcal{G}_{j}, \mathcal{G}_{j'}\}= \mathcal{F} \{\mathbf{M}_{j},\mathbf{M}_{j'}, \mathbf{A}_{j},\mathbf{A}_{j'}\}$. The literature identifies two types of constructed graphs: static and dynamic, which are described as below.

\textbf{Dynamic Graphs.} $\mathbb{G}$ is trained \textbf{\textit{dynamically}} if either of $\mathcal{G}_{j}$ or $\text{Sim} \{\cdot,\cdot\}$ has learnable parameters; e.g. an encoder network with a set of parameters $\xi_1$ can be used to provide $\mathbf{M}_{j}$ and $\mathbf{A}_{j}$ from the raw data that forms the $j$th observation \cite{dai2022graph,deng2021graph}, or a network with a set of trainable parameters $\xi_2$ can used as $\mathcal{F}$ \cite{deng2022graph,cao2022adaptive}. 

\textbf{Static Graphs.} If there are no learnable parameters assigned to $\mathcal{G}_{j}$ and/or $\text{Sim} \{\mathcal{G}_{j},\mathcal{G}_{j'}\}$, we are dealing with \textbf{\textit{static}} graphs \cite{ho2022self}. In other words, a static graph is built using a set of pre-defined node features and edges obtained from the $j$th observation of the $K$ variables. For example, with prior knowledge about the sensors' locations in a $K$-variable sensory system, a static graph at the $j$th observation is constructed where nodes are sensors and edges are defined by the Euclidean distance between sensors. Note that in some algorithms, the relations across static graphs are learned implicitly through employing a temporal-based network such as recurrent neural networks \cite{hadi2023vehicle}. Notably, \cite{hadi2023vehicle} has been shown to be the first in the G-TSAD field to incorporate a predefined $\text{Sim}\{\cdot,\cdot\}$, as an additional form of supervision within their model. This study demonstrates that utilizing $\text{Sim}\{\cdot,\cdot\}$ improves model performance compared to approaches that do not make use of this measure.

\subsubsection{Handling Non-Stationarity in Time-series Data} In many real-world applications, dynamically learning graphs is crucial since dynamic graphs allow for the adaptation of relationships between variables as the underlying data distribution changes. For example, in financial markets, the correlations between assets (variables) may fluctuate over time due to shifting market conditions. By dynamically updating the graph edges and $\text{Sim}\{\cdot,\cdot\}$ to reflect these changing relationships, dynamic graphs can effectively capture the \textbf{non-stationarity} inherent in time-series data.

\subsubsection{Addressing High Dimensionality} %As illustrated above, graphs show the ability to capture intra-variable and inter-variable dependencies, thus effectively detecting various types of anomalies within time-series data across many real-world applications. Besides, 

Graphs have also demonstrated their capability to address other challenges such as \textbf{dimensionality} \cite{khoshraftar2024survey}. One approach is to construct the static graphs based on prior knowledge \cite{ho2022self}, incorporating information such as the distance, correlations and similarity measures between variables. Based on their relevance and importance in the graph structure, graph-based methods can select features that are most discriminative or relevant for the task at hand. By focusing on a reduced set of features, dimensionality can be effectively managed while maintaining the predictive power of the data. Another strategy is to project the graph elements onto the graph embedding space using techniques such as node2vec \cite{grover2016node2vec} or graph neural networks \cite{wu2020comprehensive}, which can map high-dimensional data into a lower-dimensional space while preserving the structure and relationships encoded in the graph. Lastly, graph-based regularization techniques \cite{zhu2018robust,qu2021supervised,yang2021rethinking,ahmed2021graph} can be applied to impose constraints on the model parameters during the graph learning process. This ensures that the learned graphs exhibit desirable properties or structures of the graphs such as smoothness, where features should change smoothly between neighboring nodes, or sparsity, to prevent overly connected graphs. 

%As such, these regularization terms can mitigate the curse of dimensionality and improve generalization performance.

\subsubsection{Mitigating Noise in Time-series Data}

Graph regularization techniques can also mitigate the effect of \textbf{noise} in the data \cite{zeng2019deep,rey2021robust,li2023noise}. For example, smoothness regularization encourages gradual changes in signal values between neighboring nodes, effectively filtering out high-frequency noise components \cite{tang2023modeling}. Moreover, many methods have been developed to minimize the impact of noisy data when learning the graph structure \cite{kang2019robust,berger2020efficient,du2023noise}. These algorithms leverage robust optimization techniques to handle noise, resulting in more accurate graph representations that are less susceptible to noise.

\subsection{Graph Representations Across Different Time-Series Domains} \label{subsec:visualizations}

\begin{figure}[t]
\centering  
\includegraphics[width=8.9cm]{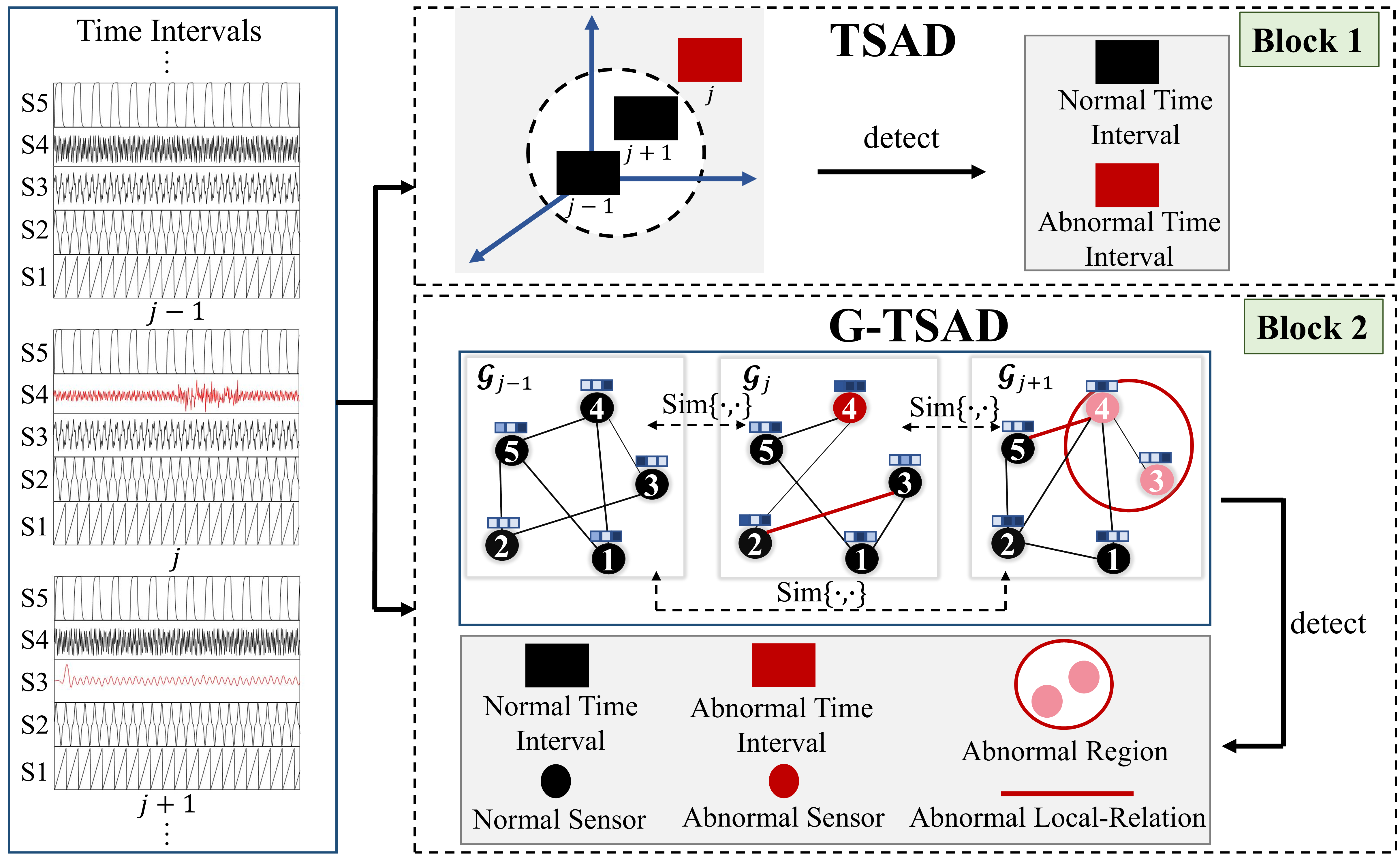}
\caption{An example of anomaly detection in a multivariate time-series signal data to show the difference between TSAD (Block 1) and G-TSAD (Block 2). The inputs are three successive time intervals (S: Sensor). In the constructed graphs, the solid and dash lines, respectively, indicate the inter-variable and intra-variable dependencies, $m = 3$, and the edge features are not shown for simplicity. Normal and abnormal cases are respectively shown in black and red colors. G-TSAD has the potential to detect anomalous sensors, local-relations, regions, and time intervals.} 
\label{fig:gtsad}
\end{figure}

Fig. \ref{fig:gtsad} visualizes G-TSAD versus the non-graph approach (i.e., TSAD).
Fig. \ref{fig:gtsad} (Left) shows raw signals from three time intervals of a multivariate system with five sensors, each recording different types of data with varying ranges and characteristics. Each sensor is then represented as a 3-dimensional feature vector, shown in both Blocks 1 and 2. Block 1 illustrates non-graph TSAD methods, where each interval is depicted as a rectangle containing data from all sensors. A key limitation of these conventional methods is their inability to explicitly model inter-sensor relationships. That is, they simply concatenate the sensor data, train the model based on this concatenated input, and aim to detect only interval-level anomalies. This diminishes their ability to detect anomalies at a finer-grained level, such as abnormal sensors, abnormal sensor interactions (aka relations), or anomalies involving multiple sensors (aka regions).

%\s{ records the car engine temperature, while the other sensor records the car speed -- so the data range and the sampling rate vary per sensor \cite{chowdhury2023primenet}.  \r{why having a different range is a change? Explain.}}}
%
%\v{\r{STILL NEEDS revision!}Fig. \ref{fig:gtsad} (Left) illustrates three raw time intervals. In the $j$th interval, Sensor \#4 exhibits a high-level abnormality (denoted in red), while in the $(j+1)$th interval, Sensor \#3 shows a low-level abnormality (shown in pink). In Block 1 of Fig. 1, each black or red rectangle represents a time interval, which concatenates data from all 5 sensors. Each sensor is represented in a 3-dimensional space with three features. Conventional TSAD methods merely detect anomalies at the interval level, identifying the entire interval as normal or abnormal. These methods overlook inter-variable dependencies, limiting their ability to detect finer-grained anomalies, such as those at the relation or region levels.}

Moreover, Fig. \ref{fig:gtsad} (Left) shows that the recording of Sensor \#4 during the $j$th interval is abnormal. While, as shown in Block 1, TSAD methods fail to identify the location of the anomaly, G-TSAD methods can localize such node anomalies by analyzing each sensor individually, shown in Block 2. In contrast, the recordings of Sensors \#3 and \#4 during the $j+1$ th interval do not exhibit significant anomalies detectable by non-graph TSAD methods. However, a well-designed G-TSAD method can identify these anomalies by considering node relationships within and between intervals through edges and $\text{Sim}\{\cdot,\cdot\}$, as discussed in Section \ref{sec:graphs}.

%\s{\r{NEEDS Revision: In many real-world applications, in addition to detecting abnormal intervals, detecting abnormal regions is very important. For example, in drug-resistant epilepsy treatment, detecting the brain's regions that are responsible of generating seizures, called Epileptogenic Zone (EZ), is critically important; this is because the current practice aims to resect EZ during the epilepsy surgery \cite{jehi2018epileptogenic}. The current practice for EZ detection is based on examining the brain signals recorded through sensors embedded within the brain (under the scalp). Considering the recorded brain signals as a sensory system suggests that an accurate region-level AD would be a great asset to neurologists when identifying EZ. As shown in Block 2 of Fig. \ref{fig:gtsad}, G-TSAD techniques are capable of detecting anomalies in a sensory system, \v{not only} at the sensor-, relation-, interval-levels, \v{but also} at the region-level due to their ability to capture both inter- and intra-variable dependencies. \v{Note that in the $(j+1)$th interval, each sensor individually appears to be normal. However, when considering their combined behavior, both Sensors \#3 and \#4 can be identified as a region-level anomaly (denoted in pink). This shows the importance of capturing inter-variable dependencies, as anomalies may not be evident when sensors are analyzed individually, but become apparent when considered collectively within a broader context.}}}

As depicted in Fig. \ref{fig:gtsad}, graphs hold potential to detect various types of anomalies in a sensory multivariate system. One may wonder whether graphs can be applied to an univariate time-series signal system, where there is only a single sensor used to record the signals, in order to detect various anomaly types such as point-level, contextual-level and trend-level anomalies defined in \cite{choi2021deep}.
%
%\s{ Note that point-level anomalies manifest as isolated data points that deviate from the surrounding data points. Contextual anomalies occur when a data point exhibits abnormal behavior relative to its context at a particular time. \s{For example, a data point that represents a temperature of 20 degree Celsius may not be abnormal on its own. However, if this temperature occurs during the winter in a region where temperatures remain below freezing, it would be deemed as a contextual anomaly due to its significant deviation from the expected behavior within the context of the season and location.} Trend-level anomalies occur when there are significant deviations from the expected trends or patterns in the time series. To this end, } 
%
Constructing graphs that represent the relationships between data points, short/long time intervals, with each denoted as a variable, can be beneficial. By analyzing the level of connectivity within the graph, anomalies can be identified as data points or time intervals that have few or weak connections to neighboring points, contexts, or patterns.

\begin{figure}[t]
\centering
\subfloat[\scriptsize{Time-Series Social Networks and Graphs (U: User)}\label{subfig:fig_user}]{\includegraphics[width=0.35\textwidth]{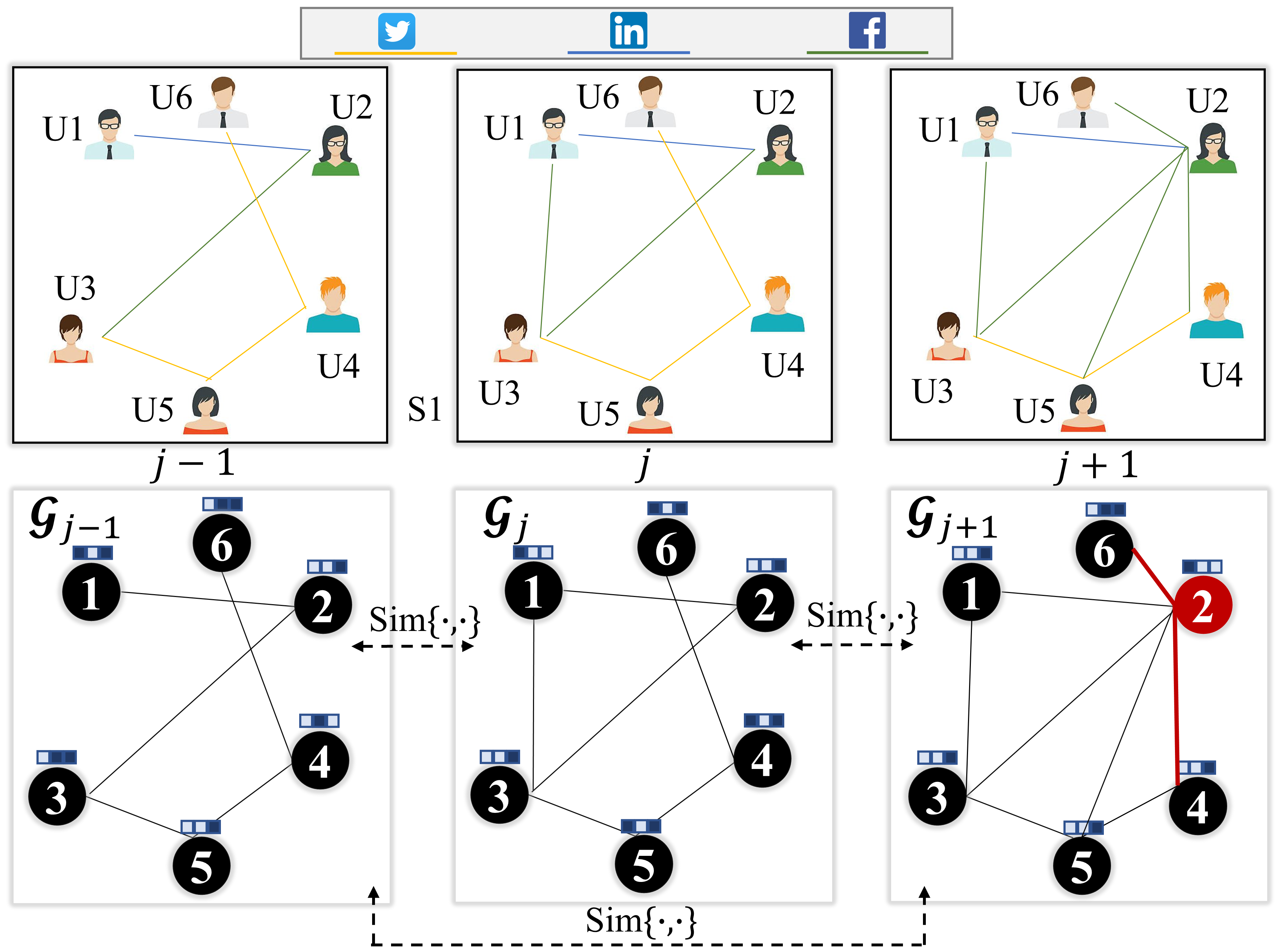}}\hspace{0.5cm}%
\subfloat[\scriptsize{Time-Series Videos and Graphs} \label{subfig:fig_video}]{\includegraphics[width=0.35\textwidth] {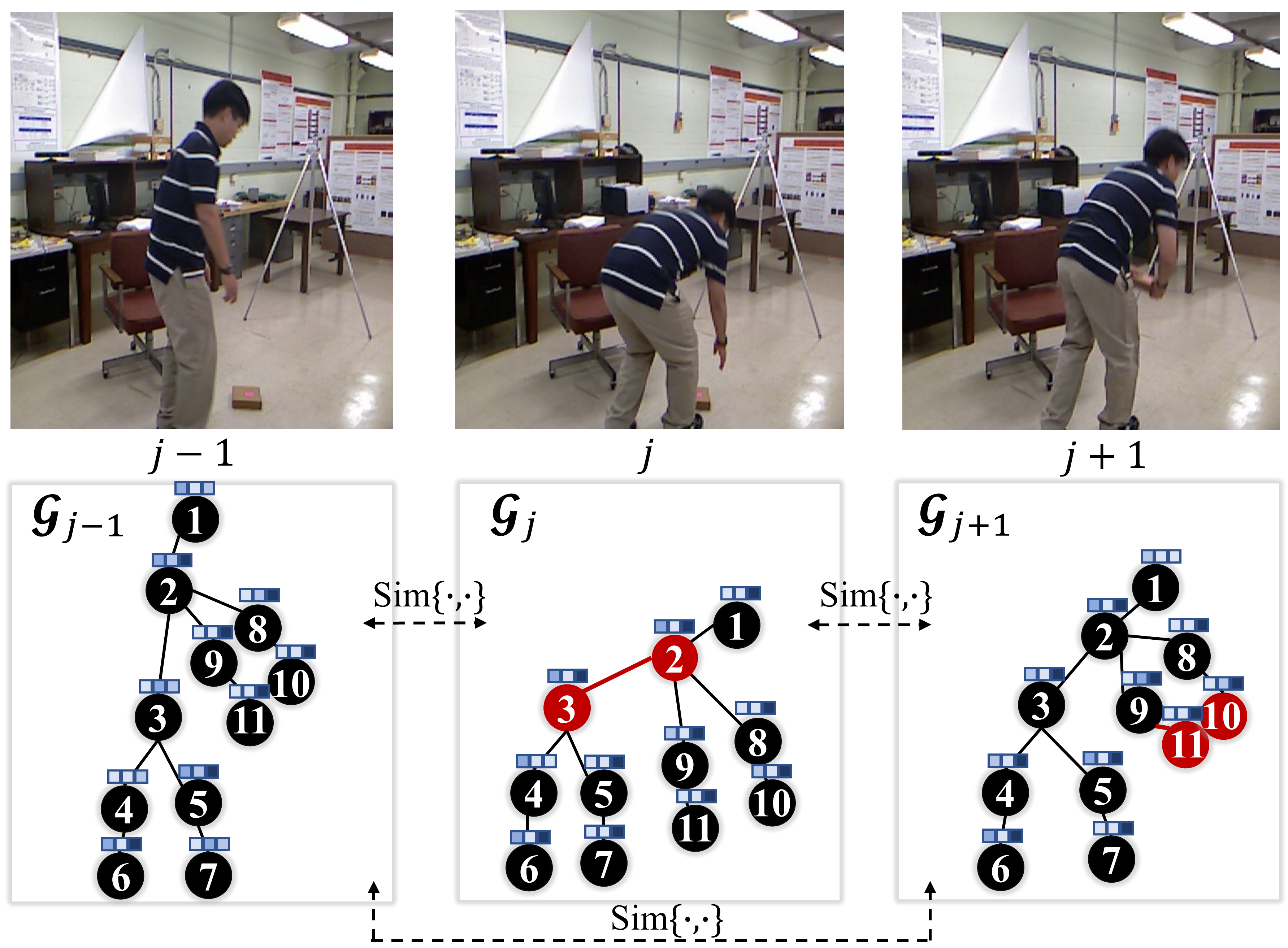}}%
\\
\caption{Examples of time-series data and the corresponding constructed graphs. Each example is shown with three consecutive observations. The top figures show the original data and the bottom figures show the constructed graphs. In the constructed graphs, the solid and dash lines, respectively, indicate the inter-variable and intra-variable dependencies, $m = 3$, and the edge features are not shown for simplicity. Normal and abnormal cases are respectively shown in black and red colors.}
\label{fig:timeseriesvsgraph}
\end{figure}

Graphs also present a capability in other time-series domains. Fig. \ref{fig:timeseriesvsgraph} shows two additional examples of time-series data and their constructed graphs by which both the intra- and inter-variable information are captured. Note that the three successive observations at the $j-1$, $j$, and $j+1$ intervals, shown in Fig. \ref{subfig:fig_user} and \ref{subfig:fig_video} respectively, correspond to three successive snapshots of a social network and three successive frames of a video. 
In social networks, the users are considered as graph nodes, and the users' interactions as edges. Most normal nodes have reasonable number of connections, which indicates the user has an ordinary social network activity. A node with an exceptionally high number of connections can be defined as an abnormal user. An abnormal node may represent a celebrity, an influential person, or a leader within the network (Fig. \ref{subfig:fig_user}). 
In video applications, a video can be modeled as a stream of time-evolving object-level graphs, where an object (e.g., a human body joint, an object in the scene) is considered as a graph node and edges represent the nodes relation within a video frame. Fig. \ref{subfig:fig_video} shows an example of anomalous activity detection where a node is assigned to every body joint -- any unexpected joint movements shall be detected as anomalies.

In summary, we have shown the capability of graphs in addressing various challenges present in time-series data described in Section \ref{sec:timeseries}. More technical insights into methods specifically tailored to address individual challenges or combinations thereof are provided in Section \ref{sec:methods}.

\section{Anomalies in Graphs} \label{sec:typesAno}

As illustrated in Section \ref{sec:graphs}, there are many types of anomalies in time-series data. For instance, anomalies may manifest as anomalous sensors, local relationships between sensors, regions, and time intervals within a sensory multivariate signal system. In a univariate time-series system, anomalies can take the form of point-level, contextual-level, and trend-level anomalies. Dynamic social networks may feature abnormal users (nodes) and relationships between users (edges) as anomalies. In video streams, unusual object movements can be regarded as anomalies. To provide a general categorization applicable to most of time-series data, we categorize anomalies that can be observed in graphs into five groups, namely anomalous nodes, edges, sub-graphs, graphs and $\text{Sim}\{\cdot,\cdot\}$. They are, respectively, known as variables, the local relations between variables, small sets of variables and their corresponding connections, the full set of variables and their corresponding connections at the $j$th observation, and the global relations between observations. Note that detecting anomalous objects in constructed graphs is much more challenging than other graph types due to the dynamic variations of nodes and edges \cite{liu2022graph}.

\begin{figure}[t]
\centering  
\includegraphics[width=8.9cm]{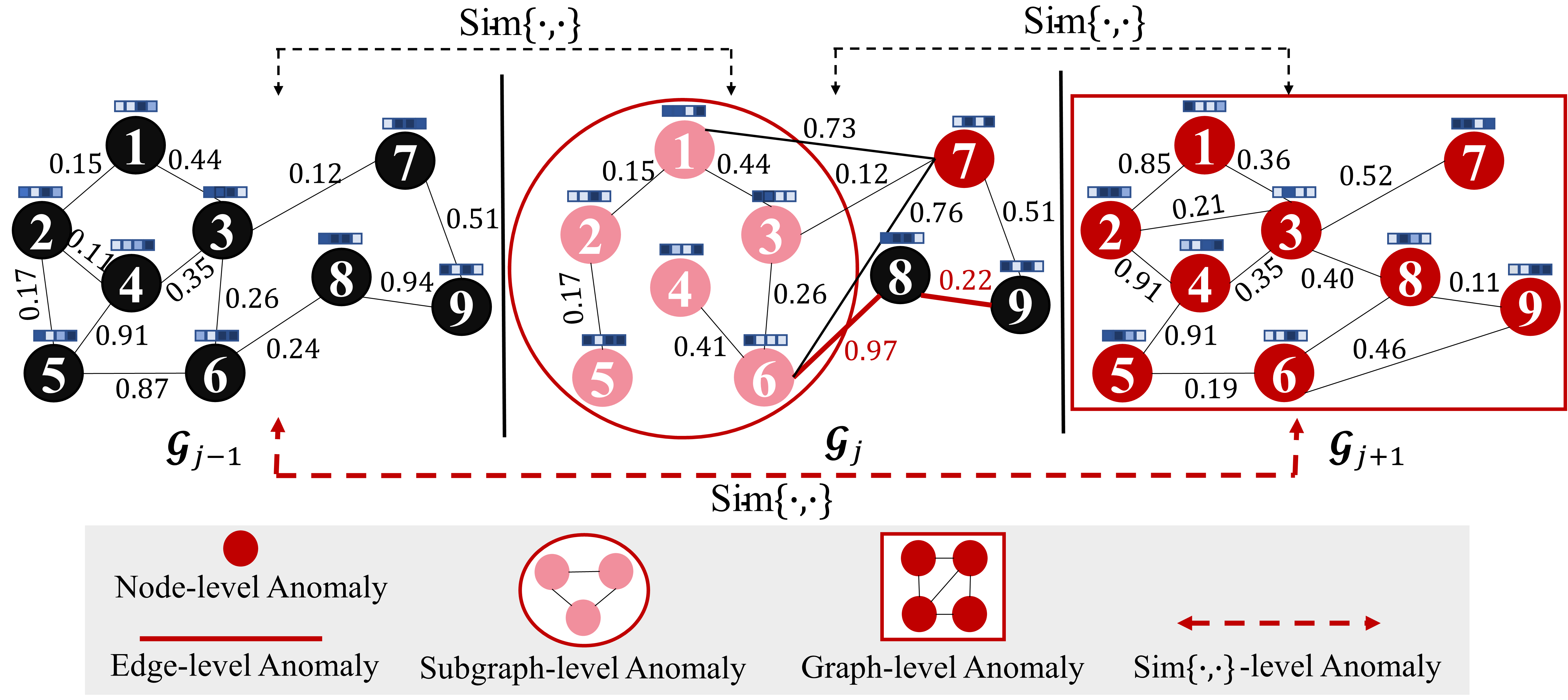}
\\
\caption{Example of node-level, edge-level, sub-graph-level, graph-level, and $\text{Sim}\{\cdot,\cdot\}$-level anomalies in a graph set $\mathbb{G}$ with three successive observations. In each graph $\mathcal{G}, m = 4$ and $m' = 1$.}
\label{fig:anomalies}
\end{figure}

Fig. \ref{fig:anomalies} shows a graph set with three successive observations, i.e. $\mathbb{G} = \Bigl\{\mathcal{G}_{j}, \text{Sim}\{\mathcal{G}_{j},\mathcal{G}_{j'}\}\Bigl\}_{j \neq j'}$.
%
%, where every $\mathcal{G}_{j}$ has nine nodes, the node's 4-dimensional feature vector ($m=4$) is depicted on top of the node, the edge's dimensional feature vector (e.g., $m'=1$) is not shown for simplicity. 
%
The five anomaly types are illustrated in the figure. In the first observation, all nodes and edges are normal. However, in the second observation, one anomalous node, two anomalous edges, and one anomalous sub-graph are appeared. 
%
%Specifically, the anomalous node \#7 and two anomalous edges (i.e., the edge between nodes \#6 and \#8, and the edge between nodes \#8 and \#9) show irregular evolution of their structures and features compared to the rest of the nodes/edges in the graph.
%
In an anomalous sub-graph, each node and edge might look normal; however, if they are considered as a group, an anomaly can be detected. As sub-graphs vary in size, as well as node and edge level properties, detecting anomalous sub-graphs is more challenging than nodes and edges.
Regarding graph-level anomalies, they are defined as abnormal graphs in a graph stream -- specifically, given a sequence of graphs, an anomalous graph at the third observation can be distinguished from other graphs based on their unusual evolving pattern of nodes, edges and corresponding features.

Lastly, anomalous $\text{Sim}\{\cdot,\cdot\}$ occurs when there are unusual relationships between graphs (aka observations). As $\text{Sim}\{\cdot,\cdot\}$ can capture both short-term and long-term relations across observations, anomalous $\text{Sim}\{\cdot,\cdot\}$ can be detected by analyzing the evolution of the graphs and their relationships over time. Assume that the three observations in Fig. \ref{fig:anomalies} correspond to the monthly stock market; short-term variations, modeled by $\text{Sim}\{\mathcal{G}_{j-1},\mathcal{G}_{j}\}$ and $\text{Sim}\{\mathcal{G}_{j},\mathcal{G}_{j+1}\}$, may represent a small price change. These small variations can be considered normal. However, a longer relationship presented in $\text{Sim}\{\mathcal{G}_{j-1},\mathcal{G}_{j+1}\}$ may raise an abnormality.

In general, given a graph $\mathbb{G}$, the goal of an ideal anomaly detector system is to produce a learnable anomaly scoring function $f(\cdot)$ that assigns an anomaly score to nodes, edges, sub-graphs, and graphs at every observation, and $\text{Sim}\{\cdot,\cdot\}$ at the coarse level. The larger $f(\cdot)$ is, the higher the probability of the graph object being abnormal. 

We introduce G-TSAD methods capable of detecting individual graph-based anomaly types or their combinations throughout Section \ref{sec:methods}. 
%
%Table \ref{tab:comparison} also lists representative methods that are able to detect types of graph-based anomalies. 
%
Note that although five types of graph anomalies are observed in the example shown in Fig. \ref{fig:anomalies}, currently, there is no study that targets detecting all types of anomalies. None of the existing methods target to detect anomalous $\text{Sim}\{\cdot,\cdot\}$ and very few works could simultaneously detect multiple anomalous objects in graphs. This will be further discussed in Section \ref{sec:discussion}.

\section{Categorization of G-TSAD Methods} \label{sec:methods}

Existing G-TSAD methods can be categorized according to several taxonomies. One taxonomy is based on the type of constructed graphs: static and dynamic, as discussed in Section \ref{subsec:constructed_graphs}. A second taxonomy considers the trainable parameter perspective, i.e., for dynamic graphs, methods can be further categorized into node, edge, and $\text{Sim}\{\cdot,\cdot\}$ learning. In the literature, no studies have explored $\text{Sim}\{\cdot,\cdot\}$ learning. Note that static graphs do not involve learnable parameters as node, edge and $\text{Sim}\{\cdot,\cdot\}$  features are predefined. A third taxonomy relates to the types of anomalies that G-TSAD methods aim to detect, including node, edge, subgraph, graph-level, and $\text{Sim}\{\cdot,\cdot\}$ anomalies. In this paper, we adopt a coarser taxonomy that encompasses all existing taxonomies mentioned above, based on the type of loss function minimized during \emph{training}. Specifically, we group methods into four categories: AE-based, GAN-based, predictive-based, and self-supervised methods. This high-level classification is elaborated in Sections \ref{sec:generative} through \ref{sec:selfsupervised}. Within each of these categories, we further provide fine-grained classifications in Tables \ref{tab:AE_methods}–\ref{tab:Selfsupervised_methods}, which summarize the type of constructed graphs (static or dynamic), learning tasks (node, edge, or $\text{Sim}\{\cdot,\cdot\}$ learning), and the types of anomalies (node, edge, subgraph, graph, or $\text{Sim}\{\cdot,\cdot\}$).

First, in AE-based methods, the reconstruction loss is the core mechanism. The model is trained to minimize the difference between the input data and its reconstruction using an encoder-decoder architecture. The goal is for the model to accurately reconstruct the normal samples, resulting in a low reconstruction error for normal data. In the test phase, anomalies are detected based on the reconstruction error. 

Second, GAN-based methods operate with a more complex mechanism, consisting of a generator and a discriminator. The generator inputs both noises and real samples to reduce the randomness issue. It aims to generate fake samples that are as realistic as the real samples, so the generator loss is the reconstruction loss to minimize the difference between the fake and real samples. Meanwhile, the discriminator uses the cross-entropy loss to classify the real samples as real and the generated samples as fake. The generator and discriminator losses together provide a richer training loss, distinguishing them from pure reconstruction-based AE methods. In the test phase, the losses of both components can be used as an indicator for anomaly scores.

Third, unlike the other categories that only consider the current and past data for loss calculation during training, predictive-based methods aim to forecast future values based on historical data. The loss function is the prediction error, which measures the difference between the predicted and actual future values. While this prediction error can be commonly viewed as a form of reconstruction error —
since the model aims to reconstruct the next time steps based on past data — the focus of predictive methods is on forecasting future values rather than reconstructing the given input data. Note that predictive-based methods, though capable of incorporating AE, GAN, or self-supervised approaches, are focused on forecasting future values rather than reconstructing input data. A sample is flagged as anomalous if the model fails to accurately predict the desired future values.

Lastly, self-supervised methods are a subgroup of unsupervised learning methods where there is no label for the training data. Compared to the three categories discussed above, self-supervised methods aim to derive more meaningful representations from the data by incorporating pretext tasks specifically designed for the unlabeled data. Their self-supervised loss functions are designed to minimize the difference between the model's predictions and the predefined outputs associated with these pretext tasks.

Note that all method categories share common elements of deep feature learning  \cite{pang2021deep}, as their training involves proxy tasks that enhance representation learning. Additionally, most of the existing methods employ a notion of encoder and decoder when developing their algorithm. Furthermore, most of existing studies train on only normal data, while the test set additionally includes anomalous data to verify the methods' detection performance. We refer to methods that only have access to normal data during training as unsupervised AD. Further details will be described in the following sub-sections. Each begins with an overview of the corresponding method category and concludes with a summary highlighting key takeaways. Accordingly, we retain the ``\textbf{Overview}" and ``\textbf{Summary}" labels to enhance readability.
%Moreover, all four categories learn graph-based representations from two perspectives: the features of nodes and edges, and the connectivity patterns between nodes represented by adjacency matrices. 

%\br{It is important to emphasize that existing surveys, such as \cite{pang2021deep}, have proposed broad taxonomies for deep learning-based AD across various data domains, including images, signals, videos, graphs, and tabular data. \cite{pang2021deep} classifies methods into generic categories: deep learning for feature extraction, learning representations of normality, and end-to-end anomaly score learning. This general framework is valuable and one could argue that AE-based, GAN-based, predictive-based, and self-supervised methods all share common elements of deep feature learning, as their training involves proxy tasks that enhance representation learning. However, our taxonomy is specifically designed for existing G-TSAD methods, where both structural (graph connectivity) and temporal (node/edge feature evolution) dependencies play a crucial role. Hence, we categorize methods based on their training loss functions, which allows for a clearer distinction between reconstruction-based, adversarial training-based, prediction-driven and self-supervised methods in the training phase. Each of these categories serves a distinct purpose in modeling graph structure and temporal dynamics in the context of G-TSAD.}

\subsection{AE-based Methods} \label{sec:generative}

\begin{figure}[t]
\centering  
\includegraphics[width=8.9cm]{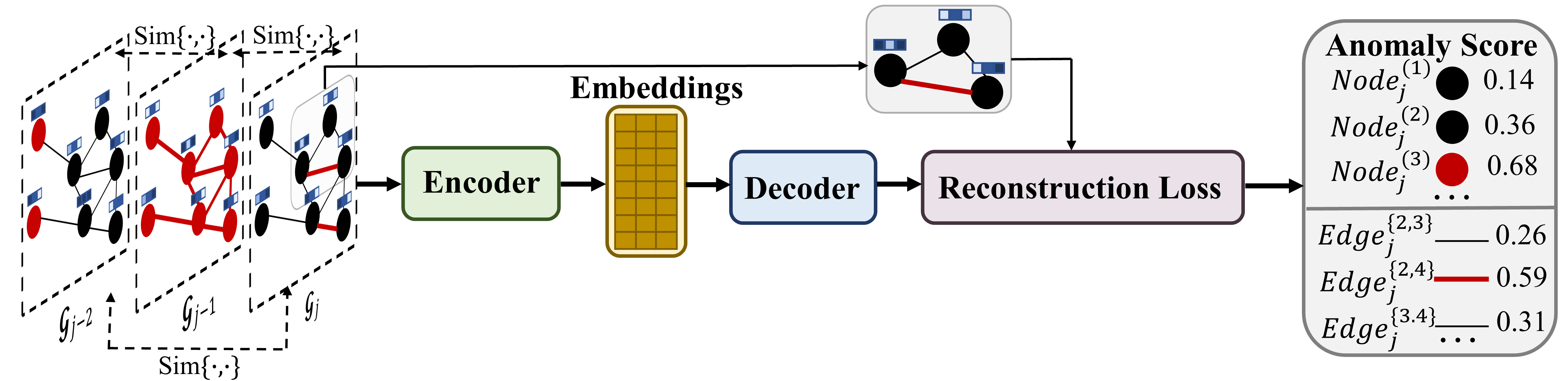}
\\
\caption{Overall framework of AE-based methods. The input is a graph $\mathbb{G}$ with three successive observations. In each $\mathcal{G}, m=3$ and the edge features are not shown for simplicity.}
\label{fig:AE}
\end{figure}

\textbf{Overview.} Fig. \ref{fig:AE} and Table. \ref{tab:AE_methods} illustrate the overall framework and representative methods in the AE-based category. In the literature, the AE-based methods take the full graph at every observation as the input and aim to reconstruct its components: nodes/edges features and the adjacency matrices. The origin of these methods can be traced back to vanilla AE, an encoder-decoder framework, where the encoder network $E_{\theta}$ (parameterized by $\theta$) learns to compress graph data into low-dimensional embeddings, and the decoder network $D_{\phi}$ (parameterized by $\phi$) aims to reconstruct the input. This can be formulated as:

\begin{equation}
    \theta^{*}, \phi^{*} = \operatorname*{argmin}_{\theta,\phi} \mathcal{L_{\text{rec}}} \Big(D_{\phi}(E_{\theta}(\mathbb{G})),\mathbb{G}\Big),
\end{equation}
where the reconstruction loss $\mathcal{L}_{\text{rec}}$ is often the mean squared error (MSE) or the cross entropy (CE) loss \cite{zhang2018generalized}.

DeepSphere \cite{teng2018deep} is the first AE-based approach for G-TSAD. It incorporates an LSTM-AE with hypersphere learning to capture both intra- and inter-variable dependencies in dynamic networked graphs. It uses adjacency matrices available in these networked graphs to reconstruct normal patterns. The idea of employing AE is that AE has a strong capability in learning non-linear patterns. Such high-quality non-linear representation supports the hypersphere learning -- the hypersphere boundary is trained to enclose the normal data in the representation space. The node-level objects lying outside the hypersphere tend to be anomalous. DeepSphere has proven effective in detecting anomalies in traffic networks and social media scenarios.

%To validate the capability of DeepSphere in learning non-linear representations of the data, we compare it with a non-AE method called MTHL \cite{teng2017anomaly}. While DeepSphere extracts non-linear features by AE, MTHL can only extract shallow features \t{in the dynamic networked graphs} by leveraging Support Vector Data Description (SVDD) \cite{tax2004support}. After having extracted shallow features, MTHL learns the hypersphere to distinguish \t{node-level} abnormal samples from normal samples \t{in the traffic and social networks}. As a result, MTHL can only achieve a suboptimal performance compared with DeepSphere as MTHL would be inefficient in AD when dealing with a complex non-linear feature space.

%Despite the great ability of DeepSphere to extract non-linear representations of data, 
%
The main challenge of this vanilla AE framework is the non-regularized latent space i.e., no specific constraints are imposed on the latent space during training. Hence, the reconstruction process is done by sampling the data points from the non-regularized latent space. This could limit the model's ability to reconstruct valid inputs in the test phase since AE may simply memorize to perfectly reconstruct the input data during training.
Variational autoencoders (VAE) \cite{kusner2017grammar} come into the picture to address this issue. VAE maps the inputs to a distribution in the latent space rather than vectors as done by the vanilla AE. This means instead of learning a single encoding for each input, VAE learns a distribution for the data. VAE imposes a constraint by which the latent space has a normal distribution; this constraint guarantees that the latent space is regularized. With such a distribution, VAE allows to represent the input with some degree of uncertainty, i.e., a more generalized data representation is learned.

Recognizing the potential of VAE, many algorithms have been developed to improve G-TSAD performance. GReLeN \cite{zhang22grelen} integrates VAE and a Graph Neural Network (GNN) with a graph structure learning for AD in multivariate time-series signals, collected from the server and water systems. Specifically, the latent space in VAE serves as the module for feature extraction, while GNN with the self-attention mechanism \cite{vaswani2017attention}, which assigns the attention weights to the edge weights in the graph, captures the between-sensor dependency. Each sensor (node) is then assigned an anomaly score based on the reconstruction error. GreLeN addresses the importance of learning the inter-variable dependency between sensors by its graph-based module, as well as the intra-variable dependency learned implicitly by its VAE. 

Deep Variational Graph Convolutional Recurrent Network (DVGCRN) \cite{chen2022deep}, another VAE-based method, was proposed to capture both intra- and inter-variable dependencies, as well as the stochasticity present in the multivariate time-series signals, collected from the server machines and satellite monitoring systems. DVGCRN includes three components: a stacked graph convolutional recurrent network to model the multilevel intra-variable dependency; a Gaussian-distributed channel embedding module to characterize the similarity and stochasticity of different sensors by computing graph edges through the inner product of sensor embeddings; and a deep embedding-guided probabilistic generative network to model the non-deterministic inter-variable dependency in the latent space; hence it can capture multilevel information at the different layers. Note that DVGCRN captures multi-level information since it extends the model into a multi-layer network. The reconstruction error is assigned for each sensor (node) as an anomaly score.

Although VAE can map the input to a distribution in the latent space to learn generalized representations of data, VAE may struggle to model high-dimensional distributions with complex structures. Thus, VAE may not reconstruct the samples that are as realistic and high-quality as the original data. Moreover, VAE models are sensitive to the hyperparameter choice and require careful tuning to achieve good performance. To tackle these issues, few recent studies, leveraging Normalizing Flow (NF) \cite{kobyzev2020normalizing,su2019robust}, have developed and yielded superior performance as NF can model complex distributions in the latent space; hence, high-quality samples can be reconstructed. NF is a statistical method using the change-of-variable law of probabilities to transform a base distribution into a target distribution. 

%OmniAnomaly \cite{su2019robust} is one of the first NF studies for G-TSAD.  OmniAnomaly incorporates a Gate Recurrent Unit (GRU) network and a VAE to capture intra-variable dependencies in the latent space. It also employs planar NF layers, which takes a series of invertible mappings to model non-Gaussian posterior distributions in the latent space to assist in learning complex intra-variable dependencies. \t{Note that OmniAnomaly implicitly represents graphs within its GRU-VAE architecture, where nodes are memory variables in GRU cells which are deterministic, and edges represent the dependence between variables.} The reconstruction score is used to rank the abnormality level. Although OmniAnomaly, with its NF property, could show the effectiveness in capturing the complex intra-variable dependency, it struggles to explicitly model between-sensor dependencies, which is crucial for multivariate sensory systems.

A recent novel NF-based method, called GANF \cite{dai2022graph} learns Directed Acyclic Graphs (DAG) inside a continuous flow model, allowing for an explicit representation of sensor dependencies in sensory systems, such as highway traffic, water systems and power grids. Specifically, in addition to learning the intra-variable dependency by an LSTM model, GANF includes a graph-based dependency encoder to capture inter-variable information in DAG. Note that DAG is a Bayesian network in which a node is conditionally independent of its non-parents. NF is included to learn DAG by modeling the conditional density of each node in DAG, i.e., NF expresses the density of each node through successive conditioning on historical data and uses conditional flows to learn each conditional density. Unlike OmniAnomaly \cite{su2019robust}, which uses the reconstruction error for anomaly scores, GANF employs density estimation, identifying nodes in low-density regions of the data distribution as anomalies.

However, there are several major issues with the NF-based methods in G-TSAD. NF can be computationally expensive, especially for modeling complex distributions for large datasets since each transformation in the flow requires computing invertibility and the determinant of the Jacobian matrix. Moreover, the resulting NF models may not be highly interpretable for high-dimensional data as it is difficult to understand the relationships between the input variables and the generated samples in such complex data.

\begin{table*}[!ht]
\caption{Summary of Representative AE-based Methods.}
\fontsize{7}{8.7}\selectfont
\centering
\begin{threeparttable}
\begin{tabular}{c|c|c|c|c|c|c|c|c}
\hline
\textbf{Method} & \textbf{Year} & \textbf{\makecell{Data \\ Type}} & \textbf{\makecell{Constructed \\ Graph}} & \textbf{\makecell{Learning \\ Task}} & \textbf{Technique}  & \textbf{\makecell{Anomaly \\ Type}} & \textbf{\makecell{Evaluation \\ Metric}} & \textbf{\makecell{Targeted \\ Application}} \\ 
 \hline
 DeepSphere \cite{teng2018deep} & 2018 & \makecell{Signals, \\ Videos} & Dynamic & \makecell{Node, \\ Edge} & \makecell{LSTM-AE with \\ hypersphare learning} & Node & \makecell{Kappa Statistics, \\ RMSE} & \makecell{Transportation, \\ Social Media} \\ \hline
 OmniAnomaly \cite{su2019robust} &  2019 & Signals &	Dynamic & \makecell{Node, \\ Edge} & \makecell{GRU-VAE with \\ planar NF layers} & Node & Prec, Rec, F1 & \makecell{Server Machines, \\ Satellite Monitoring} \\ \hline
 DVGCRN \cite{chen2022deep} & 2022 & Signals & Dynamic & \makecell{Node, \\ Edge} & \makecell{GNN-VAE with a Gaussian-\\distributed channel module} & Node & Prec, Rec, F1 & \makecell{Server Machines, \\ Satellite Monitoring}  \\ \hline
 GReLeN \cite{zhang22grelen} &  2022 & Signals & Dynamic & \makecell{Node, \\ Edge} & \makecell{GNN-VAE with a \\ self-attention module} & Node & Prec, Rec, F1 &	\makecell{Water Systems, \\ Server Machines} \\ \hline
 GANF \cite{dai2022graph} &  2022 & Signals & Dynamic & \makecell{Node, \\ Edge} & \makecell{LSTM-NF with \\ Directed Acyclic Graphs} & Node & AUC &	\makecell{Highway Traffic, \\ Power Grids} \\ \hline
 CAN \cite{xia2023coupled} &  2023 & Signals & Dynamic & \makecell{Node, \\ Edge} & \makecell{GNN with a \\ couple attention module} & Graph & Prec, Rec, F1 & \makecell{Water, \\ Soil Moisture Systems}  \\ \hline
 TSAD-C \cite{ho2023multivariate} &  2024 & Signals & Dynamic & \makecell{Node, \\ Edge} & \makecell{Diffusion models with \\ a graph modeling module} & Graph & F1, Rec, APR & \makecell{Server Machines, \\ Biomedical Systems} \\ \hline 

\end{tabular}
% \begin{tablenotes}
%   \fontsize{7}{10}\selectfont
%   \item $^{*}$ \footnotesize{App.: Application.}
% \end{tablenotes}
\end{threeparttable}
\label{tab:AE_methods}
\end{table*}

The above unsupervised methods assume that the training data must be clean. However, in real-world applications, collecting clean data poses significant challenges due to its time-consuming, costly, and labor-intensive nature. Moreover, anomalies often sneak into normal data, regarded as contaminated data, that come from the data shift or human error \cite{jiang2022softpatch}. Consequently, existing unsupervised methods, which extensively train on normal data to model its behavior, would misdetect anomaly samples encountered during training in the test phase. To address such challenges, \cite{ho2023multivariate} introduced an approach, called TSAD-C. TSAD-C comprises three modules: a Decontaminator, aimed at eliminating abnormal patterns; a Long-range Variable Dependency Modeling, designed to capture long-range intra- and inter-variable dependencies, where graphs are learned through a self-attention mechanism with the attention weights assigned for the edges' weights; and Anomaly Scoring based on reconstruction scores used to detect graph-level anomalies in both industrial and biomedical multivariate signal domains.

Note that TSAD-C employs a diffusion model - a type of generative models \cite{tashiro2021csdi,rasul2021autoregressive}, in its Decontaminator. However, similar to the drawbacks of NF, training a diffusion model can be computationally expensive, especially for large datasets. It also demands large memory resources during training when storing intermediate data representations.  Moreover, diffusion models usually rely on various hyperparameters, including learning rates, batch sizes, and model architecture, which can significantly influence performance. Effectively tuning these hyperparameters can be challenging and time-consuming, requiring extensive experimentation.

\textbf{Summary.} Many studies in the AE-based methods, ranging from vanilla AE, VAE to NF, have shown promising results in G-TSAD. However, designing a model that can capture both intra- and inter-variable dependencies requires careful consideration of multiple factors. Large amounts of data and computational resources for training are required to produce satisfactory results, which can be expensive and time-consuming for many time-series applications. Also, interpretability are necessary to understand the model's behavior with the detection outcomes. This ensures that the model makes unbiased decisions on detecting anomalies. Moreover, most of the AE-based methods limit their applications to only time-series signal systems while there are many other data types such as videos, social networks, and edge streams, which are common in real-world scenarios. This restricts the adaptability and applicability of AE-based approaches to a wider variety of time-series contexts.

\subsection{GAN-based Methods} \label{sec:gan}

\begin{figure}[t]
\centering  
\includegraphics[width=8.9cm]{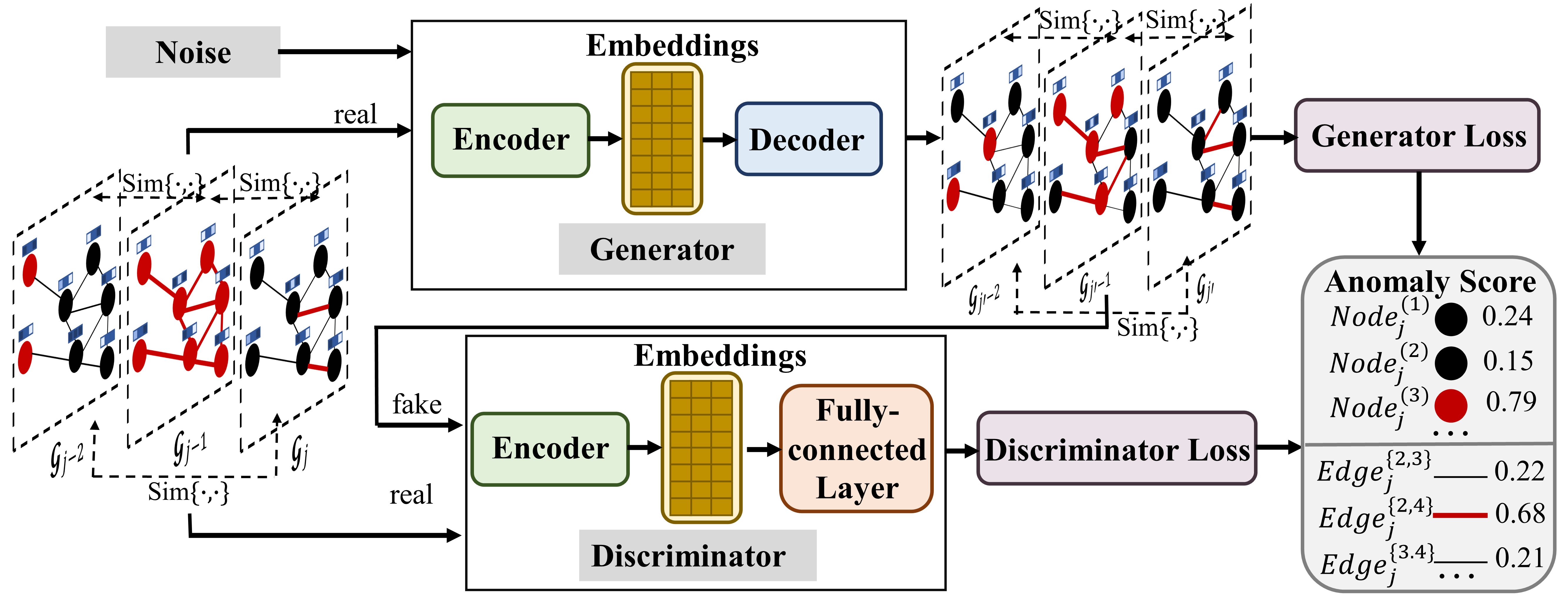}
\\
\caption{Overall framework of GAN-based methods. The input is a graph $\mathbb{G}$ with three successive observations. In each $\mathcal{G}, m=3$ and the edge features are not shown for simplicity.}
\label{fig:GAN}
\end{figure}

\textbf{Overview.} Fig. \ref{fig:GAN} and Table. \ref{tab:GAN_methods} provide the concept map and representative methods in the GAN-based category. 
%
%GAN has yielded state-of-the-art performance in many domain applications as GAN, despite the VAE and NF-based methods, can reconstruct high-quality samples without the need for explicit estimation of the distribution \cite{goodfellow2020generative}.
%
Overall, parameters of the two components of GAN, i.e. the generator and the discriminator, can be found as below:

\begin{align}
    \theta^{*}, \phi^{*} &= \operatorname*{argmin}_{\theta,\phi} \mathcal{L}_{\text{gen}} \Big(D_{\phi}(E_{\theta}(\mathbb{G}, z)),\mathbb{G}\Big), \\
    \psi^{*} &= \operatorname*{argmin}_{\psi} \mathcal{L}_{\text{disc}} \Big(\mathbb{D}_{\psi} \Big(\mathbb{G'},\mathbb{G}\Big)\Big),
\end{align}
where $z$ is the random noise vector, $\mathbb{D}$ is the discriminator network (parameterized by $\psi$), $\mathbb{G'}$ is the fake set of graph samples generated by the generator. The generator loss $\mathcal{L}_{\text{gen}}$ is often MSE, and the discriminator loss $\mathcal{L}_{\text{disc}}$ is often CE.

Recent studies have applied GAN to G-TSAD. For example, \cite{liang2021consistent} proposed Cross-Correlation Graph-Based Encoder–Decoder GAN (CCG-EDGAN) for unsupervised AD in multivariate signals, collected from satellite, wind turbine monitoring, and power consumption systems. CCG-EDGAN consists of three modules: a preprocessing module to transform signals into predefined correlation graphs; a generator to generate the fake data via the encoder-decoder-encoder structure, which captures the features in the correlation graphs; and a discriminator to classify the real from reconstructed data. Note that the input includes both noise and real data to help the generator improve the reconstructed samples' diversity and avoid the mode collapse problem, which refers to the situation that the generator can only produce the fake samples that look similar to each other but are not diverse enough to represent the full range of possible samples. The anomaly score assigned for each node is computed by the reconstruction error.

CCG-EDGAN encounters two limitations. First, it could only handle the inter-variable dependency through the correlation graphs, while the intra-variable dependency could not be captured by the GAN module. Second, its anomaly score computation is only based on the reconstruction error obtained by the generator, ignoring the discriminator's potential in detecting anomalies. To overcome the first limitation, HAD-MDGAT was proposed \cite{zhou2022hybrid}. Unlike CCG-EDGAN, which relies on predefined correlation graphs, HAD-MDGAT uses a graph attention network to learn graphs dynamically from attention weights, allowing it to capture inter-sensor correlations. A GRU network is used to capture the intra-variable patterns. Node-level anomalies are detected by a reconstruction scheme via GAN. HAD-MDGAT also compares the performance between GAN and vanilla AE and confirms that GAN achieves a better reconstruction and could solve the overfitting problem of vanilla AE. Despite the ability to tackle the first limitation of CCG-EDGAN, HAD-MDGAT could not tackle the second issue, i.e., HAD-MDGAT only uses the anomaly scores obtained by the generator. HAD-MDGAT shows its applications in satellite monitoring and soil moisture signal systems.

Several recent methods address both limitations, i.e. capturing both intra- and inter-variable information, as well as using both the generator and discriminator for computing the anomaly score. STGAN \cite{deng2022graph} was proposed to detect node-level anomalies in the traffic networks. STGAN first constructs graphs for the traffic network, with edge weights calculated using the Euclidean distance between nodes. STGAN then includes two components: a generator to model intra- and inter-variable patterns, trend, and external features; and a discriminator to distinguish the real samples from the fake samples. STGAN calls its components, respectively, spatiotemporal generator and spatiotemporal discriminator since both are based on graph convolutional gated recurrent unit that explores the intra- and inter-variable correlations among neighboring nodes. Anomaly score is computed based on the performance of both the generator and discriminator. While the generator detects sudden changes in traffic networks, the discriminator detects traffic anomalies such as abnormal times and locations.

\begin{table*}%[!ht]
\caption{\br{Summary of Representative GAN-based Methods.}}
\fontsize{7}{8.7}\selectfont
\centering
\begin{threeparttable}
\begin{tabular}{c|c|c|c|c|c|c|c|c}
\hline
\textbf{Method} & \textbf{Year} & \textbf{\makecell{Data \\ Type}} & \textbf{\makecell{Constructed \\ Graph}} & \textbf{\makecell{Learning \\ Task}} & \textbf{Technique} & \textbf{\makecell{Anomaly \\ Type}}& \textbf{\makecell{Evaluation \\ Metric}} & \textbf{\makecell{Targeted \\ Application}} \\ 
 \hline
 CCG-EDGAN \cite{liang2021consistent} &  2021 & Signals & Static & -- & \makecell{GNN-GAN with \\ cross-correlation graphs} & Node & Prec, Rec, F1 &	 \makecell{Wind Turbine, \\ Power Consumption}  \\  \hline
 HAD-MDGAT \cite{zhou2022hybrid} &  2022 & Signals & Dynamic & \makecell{Node, \\ Edge} & \makecell{GNN-GAN with an \\ attention module} & Node & Prec, Rec, F1 & \makecell{Satellite and \\ Soil Moisture Systems} \\  \hline
 STGAN \cite{deng2022graph} &  2022 & Signals & Static & -- & \makecell{GAN with both spatiotemporal \\ generator and discriminator} & Node & Rec &	Traffic Data \\  \hline
 RegraphGAN \cite{guo2023regraphgan} &  2023 & \makecell{Social \\ Networks} & Dynamic &	Edge & \makecell{Reverse GAN with edge-based \\ substructure sampling} & Edge & AUC & \makecell{Social Networks, \\ Cybersecurity} \\  \hline

\end{tabular}
% \begin{tablenotes}
%   \fontsize{7}{10}\selectfont
%   \item $^{*}$ \footnotesize{App.: Application.}
% \end{tablenotes}
\end{threeparttable}
\label{tab:GAN_methods}
\end{table*}

While HAD-MDGAT can overcome the first challenge and STGAN can tackle both challenges seen in CCG-EDGAN, the major issue of both HAD-MDGAT and STGAN methods is that, unlike CCG-EDGAN that uses both noises and the real data as the input to improve the generated samples diversity and unlike any GAN-based methods applied on images that use only noises as the input, HAD-MDGAT and STGAN set only the real data as the input and no noise is used to generate fake samples. This potentially leads to the mode collapse problem in the GAN framework.

\textbf{Summary.} Although GAN-based methods have made important contributions to the G-TSAD field, they still have some challenges that require further research. Many GAN-based methods suffer from the mode collapse problem when the generator can not produce a wide range of samples since taking only the real data as the initial input can make the generator create the same or very similar samples. Moreover, as the generator and the discriminator constantly compete against each other, the training process could be unstable and slow. Also, GAN-based methods are sensitive to the choice of hyperparameters; hence, carefully choosing them to avoid unstable training, mode collapse, and poor-quality generated samples is needed. Importantly, it is essential to design a GAN that can properly combine the detection ability of both the generator and discriminator. Lastly, to date, there are only a few GAN-based methods applied to G-TSAD, and their applicability is largely limited to time-series signal data from traffic or sensory systems. Exploring effective GAN-based methods that can be adapted to other data types, such as videos and dynamic social networks, would be a valuable direction for future research.

\subsection{Predictive-based Methods} \label{predictive}

\begin{figure}[t]
\centering  
\includegraphics[width=8.9cm]{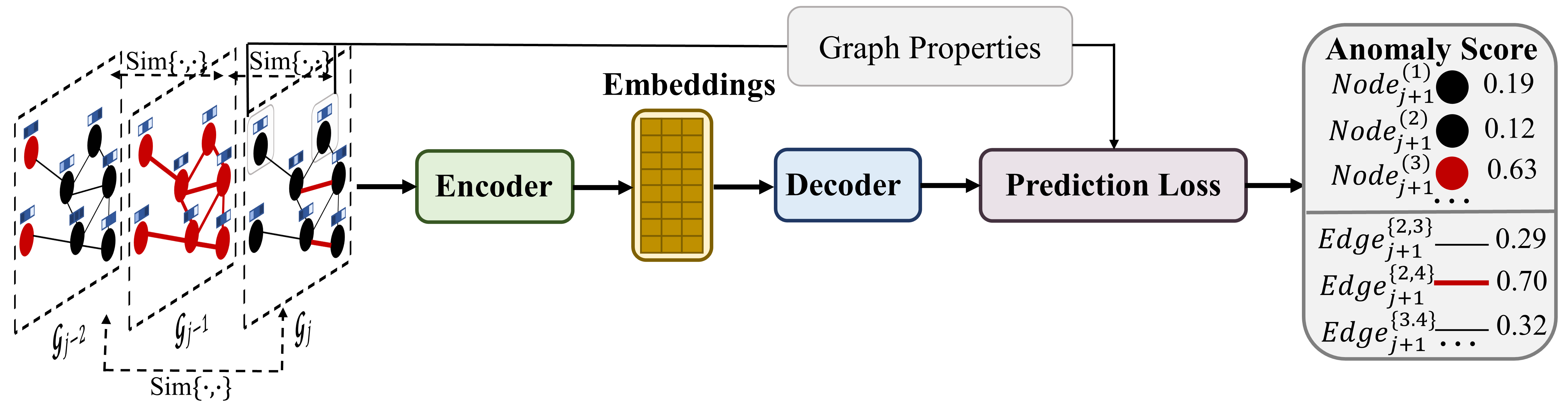}
\\
\caption{Overall framework of Predictive-based methods. The input is a graph $\mathbb{G}$ with three successive observations. In each $\mathcal{G}, m=3$ and the edge features are not shown for simplicity.}
\label{fig:predictive}
\end{figure}

\textbf{Overview.} Fig. \ref{fig:predictive} and Table. \ref{tab:Predictive_methods} illustrate  the overall framework and representative methods in the predictive-based category. As mentioned earlier, predictive methods are different from other categories as they aim to predict future values based on current and past values. A sample is detected as an anomaly if the model can not provide accurate time-series forecasting. Specifically, at each observation $j$, a predictive-based approach calculates an expected graph $\bar{\mathcal{G}}_{j+1}$. The anomaly score is then computed as the difference between the expected graph $\bar{\mathcal{G}}_{j+1}$ and the actual graph $\mathcal{G}_{j+1}$. The time-series prediction problem is difficult, as learning the long-distance intra-variable complexity of observations is necessary. As a result, in addition to the features of nodes/edges, and the connectivity patterns between nodes represented by the adjacency matrices, other important properties, such as the node degree, shall be considered -- this enables better capturing of the long-distance dependencies. This framework can be defined as follows:
\begin{equation}
    \theta^{*}, \phi^{*} = \operatorname*{argmin}_{\theta,\phi} \mathcal{L}_{\text{pred}} \Big(D_{\phi}(E_{\theta}(\mathbb{G})),\mathcal{G}_{j+1}\Big),
\end{equation}
where the prediction loss $\mathcal{L}_{\text{pred}}$ is often MSE, Root MSE (RMSE), or CE. The first term in the loss represents $\bar{\mathcal{G}}_{j+1}$.

Many recent studies have shown the potential of GNN with a graph attention network in detecting anomalies in the future. For example, \cite{deng2021graph} proposed Graph Attention-based Forecasting (GDN) to dynamically learn the graph by an attention mechanism with attention weights assigned for edge weights, then predict the behavior of each sensor at each time step by learning the past. This helps detect sensor (node) anomalies that deviate from their expected behavior. An anomaly score for each sensor at each time step is computed using the absolute error value. Normalization is then applied to all anomaly scores across all time steps for each sensor. Importantly, the obtained attention weights are used to provide explainability for the detected anomalies. GDN addresses very important issues in G-TSAD research, i.e different sensors may have different behaviors and properties (e.g., some sensors measure the water pressure while others measure the flow rate). However, existing GNN-based methods use the same set of model parameters for all sensors. Moreover, many of them pre-define the graph nodes/edges/adjacency matrices. Though using the pre-defined matrices results in a more stable training phase, defining the matrices themselves requires prior knowledge about the dataset, which may not be reachable in many applications. In spite of the GDN's great capability to address real-world issues, its limitation is that it detects anomalies mainly based on inter-variable dependencies; the intra-variable relations across different time steps are ignored.

%Inspired by GDN, FuSAGNet \cite{han2022learning} also proposes to use a GNN and a graph attention network to predict the future time-series behavior \t{in water systems and security applications. Similar to GDN, FuSAGNet also learns the graph through a self-attention mechanism.} In addition to the prediction task as of GDN, FuSAGNet includes the reconstruction task through using a Sparse-AE that induces sparsity in its latent space. In the end, a sparsity-constrained joint optimization is introduced to optimize both Sparse-AE and GDN components. Thus, FuSAGNet can achieve more accurate anomaly scores than those obtained from standalone GDN, as FuSAGNet computes anomaly scores from both the reconstruction and forecasting modules. However, similar to the challenge seen in the standalone GDN, FuSAGNet only concentrates on capturing the inter-variable dependency, and the intra-variable dependency remains unaddressed.

To overcome the limitations of GDN, MTAD-GAT \cite{zhao2020multivariate} was proposed. It leverages the graph attention network for learning the graphs and capturing both inter-, and intra-variable dependencies. The key ingredient in MTAD-GAT is that there are two graph attention layers with respect to the inter- and intra-variable information: the feature-oriented graph attention layer to capture the relations between sensors and the time-oriented graph attention layer to capture the dependency among observations. MTAD-GAT jointly trains a forecasting-based module, for single-observation prediction with RMSE, and a reconstruction-based VAE module, for capturing the data distribution of the entire time series. Although MTAD-GAT could address the issue in GDN studies, its graph attention module could not provide interpretability on the characteristics of different sensor types, present in the multivariate sensory systems, as done by other GDN works \cite{deng2021graph,han2022learning}.

\begin{table*}%[!ht]
\caption{Summary of Representative Predictive-based Methods.}
\fontsize{7}{8.7}\selectfont
\centering
\begin{threeparttable}
\begin{tabular}{c|c|c|c|c|c|c|c|c}
\hline
\textbf{Method} & \textbf{Year} & \textbf{\makecell{Data \\ Type}} & \textbf{\makecell{Constructed \\ Graph}} & \textbf{\makecell{Learning \\ Task}} & \textbf{Technique} & \textbf{\makecell{Anomaly \\ Type}} & \textbf{\makecell{Evaluation \\ Metric}} & \textbf{\makecell{Targeted \\ Application}} \\ 
 \hline

 SEDANSPOT \cite{eswaran2018sedanspot}  & 2018 & \makecell{Edge \\ Streams} & Dynamic & Edge & \makecell{Rate-adjusted sampling and \\ random walked edge techniques}  & Edge & Prec, Rec, F1 &	\makecell{Traffic Networks, \\ Social Media} \\ \hline
 GCLNC \cite{zhong2019graph} &	 2019 & Videos & Dynamic & \makecell{Node, \\ Edge} & \makecell{A label noise cleaner  with GNN \\ and a supervised action classifier} & Node & AUC &	Real-world Videos \\ \hline
 MTAD-GAT \cite{zhao2020multivariate} &  2020 & Signals & Dynamic & \makecell{Node, \\ Edge} & \makecell{A forecasting-based module, \\ and a reconstruction-based VAE} & Node & Prec, Rec, F1 & \makecell{Satellite Monitoring, \\ Server Systems} \\ \hline
 Midas \cite{bhatia2020midas} &  2020 & \makecell{Edge \\ Streams} & Dynamic & Edge & \makecell{An online method to process each \\ edge
in constant time and memory} & Edge & AUC & \makecell{Cybersecurity, \\ Social Media} \\ \hline
 GDN \cite{deng2021graph} &  2021 & Signals & Dynamic & \makecell{Node, \\ Edge} & \makecell{GNN with an attention \\ mechanism} & Node & Prec, Rec, F1 &	Water Systems \\ \hline
 GTA \cite{chen2021learning} &  2021 & Signals & Dynamic & \makecell{Node, \\ Edge} & \makecell{Gumbel-softmax sampling for \\ connection learning policy} & Graph & Prec, Rec, F1 & \makecell{Water Systems, \\ Satellite Monitoring} \\ \hline
 Eland \cite{zhao2021action} &  2021 & \makecell{Social \\ Networks} & Dynamic & Edge & \makecell{An action predictor with \\ sequence augmentation techniques} & Graph & Prec, AUC &	Social Networks \\ \hline
 Series2Graph \cite{boniol2022series2graph} &  2022 & Signals & Dynamic & \makecell{Node, \\ Edge} & \makecell{A two-embedding space \\ for the densest regions in graphs} & Sub-graph & Accuracy & \makecell{Various \\ Applications} \\ \hline
 WAGCN \cite{cao2022adaptive} &  2022 & Videos & Dynamic & \makecell{Node, \\ Edge} & \makecell{GNN with a graph \\ construction module} & Graph & AUC &	Real-world Videos \\ \hline
 FuSAGNet \cite{han2022learning} &  2022 & Signals & Dynamic & \makecell{Node, \\ Edge} & \makecell{Sparse-AE and GNN with \\ an attention mechanism} & Node & Prec, Rec, F1 & \makecell{Water Systems, \\ System Security} \\ \hline
 MST-GAT \cite{ding2023mst} &  2023 & Signals & Dynamic & \makecell{Node, \\ Edge} & \makecell{A multimodal GNN with a \\ graph attention module} & Node & Prec, Rec, F1 & \makecell{Various Industrial \\ Applications} \\ \hline
 GiCiSAD \cite{gicisadwacv2024} &  2025 & Videos & Dynamic & \makecell{Node, \\ Edge} & \makecell{Diffusion models with a \\ GNN-based forecasting module} & Graph & AUC &	\makecell{Human Activity \\ Recognition} \\ \hline

\end{tabular}
% \begin{tablenotes}
%   \fontsize{7}{10}\selectfont
%   \item $^{*}$ \footnotesize{App.: Application.}
% \end{tablenotes}
\end{threeparttable}
\label{tab:Predictive_methods}
\end{table*}

As seen from the above studies, a graph structure learning strategy (i.e., learning the features of nodes/edges and adjacency matrices) via attention mechanisms is important and commonly used in many time-series applications. This is because, with limited prior knowledge, it is difficult to pre-define the graph structure that can carry comprehensive information about the graph. Meanwhile, uncovering the important patterns in graphs is possible through learning with GNN. However, many existing methods learn the adjacency matrices in the sensor embedding space by defining measurement metrics such as the cosine similarity or top-k nearest neighbors (kNN) \cite{zhu2021survey}. Cosine similarity, which computes the dot product of all sensor embeddings, results in a fully connected graph, leading to quadratic time and space complexity with respect to the number of sensors. top-kNN is another approach based on a similarity metric, such as cosine similarity, Gaussian similarity, or Euclidean distance, that selects only the top-k neighboring nodes, which can not entirely underline the strong and wide-spread connections among sensor embeddings.

To overcome this challenge, GTA \cite{chen2021learning} presents the connection learning policy based on the Gumbel-softmax sampling approach. This is to learn bi-directed edges between sensors, thereby avoiding the issue of selecting node neighbors by the top-kNN method. GTA also handles the intra-variable dependency from the graph sequence by a transformer-based graph convolution network. Moreover, to tackle the quadratic complexity, GTA proposes a multi-branch attention mechanism to substitute the original-head self-attention method. A forecasting-based strategy is then adopted to predict the graph at the next time step and return an anomaly score assigned for each graph by MSE. GTA demonstrates its effectiveness in multivariate sensory systems, such as water and satellite monitoring systems.

In contrast to detecting abnormal sensors (nodes) of the graph in many multivariate signal systems, several studies show an interest in detecting abnormal edges (e.g., the unexpected interactions between sensors or the anomalous edges in an edge stream). SEDANSPOT \cite{eswaran2018sedanspot} considers an AD problem in an edge stream, where the edge anomalies tend to connect parts of the graph that are sparsely connected or occur as bursts of activity. An example of a burst is an occasion (e.g., a burst of longer-than-usual phone calls during festivals). SEDANSPOT exploits these abnormal behaviors by proposing a rate-adjusted sampling technique that downsamples edges from bursty periods of time and employs a holistic random walk-based edge anomaly scoring function to compare incoming edges and the whole graph. This approach is applied in traffic networks, co-authorship analysis, and social media scenarios.

Midas \cite{bhatia2020midas} also tackles the same problem in detecting anomalous edges in an edge stream, particularly for cybersecurity and social media applications. However, unlike SEDANSPOT, which identifies individual abnormal edges, Midas aims to detect microcluster anomalies, which are the sudden appearing bursts of activity sharing many repeated edges. Midas provides explanations on the false positive rate and is also an online method that processes each edge in constant time and memory, while SEDANSPOT does not address these issues. Moreover, SEDANSPOT and Midas propose different hypotheses on anomaly scores. SEDANSPOT designs an anomaly scoring function that gives a higher score to an edge if adding it to a sample of edges produces a large change in distance between its incident nodes. Midas computes the Gaussian likelihood of the number of edge occurrences in the current timestamp and declares an anomaly if the likelihood is below an adjustable threshold, which is to guarantee low false positive probability.

Not limited to node and edge AD in the context of time-series data, several predictive-based studies show the interesting task of detecting anomalous sub-graphs. As mentioned in Section \ref{sec:typesAno}, detecting abnormal sub-graphs is much more difficult than abnormal nodes and edges since abnormal sub-graphs vary in size and inner structures. It is even more difficult when individual nodes/edges are normal but abnormal when considered as a group of nodes/edges.

Series2Graph \cite{boniol2022series2graph} tackles these problems by proposing a novel idea for detecting anomalous sub-graphs with varying sizes in the graphs, which are constructed from multivariate time-series signals, collected from both industrial and biomedical domains. Specifically, Series2Graph first extracts sub-sequences (aka time intervals) from the time series using a sliding window technique. It then projects the sub-sequences onto a two-dimensional embedding space. The graph is constructed by considering the densest regions in the two-dimensional embedding space, where a notable concentration of projected sub-sequences is observed. These dense regions are selected as nodes, which serve as summaries of the repetitive patterns observed in the data, indicating normal behavior. To create the edges, Series2Graph determines the edge weights based on the frequency of occurrences of one sequence immediately following another in the input time series. If there is a transition between two nodes, they are connected by an edge; otherwise, no edge is created. Series2Graph determines the abnormality level of the sub-graphs using the node degree (aka the number of connected edges to the node) and the edge weights (aka the number of consecutive occurrences of two sequences).

%Another interesting sequence prediction study, namely Eland \cite{zhao2021action}, proposes a sequence augmentation technique for early-stage graph AD when the data is incomplete. Unlike Series2Graph, which first converts sequences into graphs and targets to identify abnormal sub-graphs, Eland converts the original graphs of social networks to sequences and aims to detect abnormal \t{graph-level} sequences. Eland develops a sequence-based action predictor (i.e., Seq2Seq) to forecast the users' behaviors based on their activity history. Eland uses a GNN to train augmented graph-based sequences with the CE loss to minimize the difference between predicted and actual sequences. While Eland demonstrates that its sequence augmentation technique yields improvements on AD, it only aims to analyze social network data. Dealing with other types of time-series data, such as time-series signals, which are noisy, complex, and non-stationary in nature, would bring a big challenge for Eland.

Existing predictive-based methods have shown the capability to deal with many time-series applications, ranging from time-series signals, edge streams to social networks. However, none of these approaches are applicable to videos in which a video is a sequence of image frames indexed in time. GCLNC \cite{zhong2019graph}, WAGCN \cite{cao2022adaptive}, and WSANV \cite{li2022weakly} are some of the earliest G-TSAD studies working on real-world surveillance videos. While most of the existing studies use only normal data during their training phase, these three studies have taken advantage of weakly-supervised labels (e.g., include noisy labels as wrong annotations of normal graphs) during training to improve the detection of abnormal graph-level frames in videos. As a video sequence has temporal evolution between frames, these methods construct graphs, where each frame is modeled as a graph node, and the adjacency matrices are computed based on not only the similarity metric of the inter-variable features but also the intra-variable proximity, i.e., the short-distance intra-variable dependency to capture relationships among frames. Note that instead of pre-defining the adjacency matrices, they adjust them as the model is trained. A computed anomaly score based on CE is then assigned for each frame. 

Despite the capability of these methods on videos, they only consider frame-level graphs, hence, the relationships between frames are captured. However, in practice, it is essential to model a video as a stream of time-evolving object-level graphs, where a node is an object and edges represent the relationships among objects. Object-level graphs can help to determine the abnormal changes in actions/movements of an object across frames in a video.

\textbf{Summary.} Predictive-based methods have shown great potential in various time-series applications, from time-series signals, edge streams, social networks to videos. Interestingly, many predictive-based methods integrate the reconstruction and prediction tasks to yield more accurate anomaly scores. However, several challenges still remain unsolved. While existing methods have shown their effectiveness in building graphs to capture inter-variable information, one of the biggest challenges of these methods is modeling the long-term intra-variable dependency, which is required to effectively perform time-series forecasting tasks \cite{chen2023long}. For example, existing methods only consider the intra-variable dependency in a short-term period where observations close to each other in the time series are expected to be similar. However, as mentioned in Section \ref{sec:timeseries}, trend, seasonality, and unpredictability patterns are, in nature, always available in time-series data. Hence, forecasting future values simply based on near-distance past values is problematic. Plus, a large amount of data with a diversity of patterns is required during training for the forecasting tasks, but in practice, having access to a comprehensive dataset is very difficult in many time-series applications. For example, in medical systems, it is difficult to obtain sufficient patient data that covers all types of diseases and rare conditions due to privacy concerns and limited access to patients over time.

\subsection{Self-supervised Methods} \label{sec:selfsupervised}

\begin{figure}[t]
\centering  
\includegraphics[width=8.9cm]{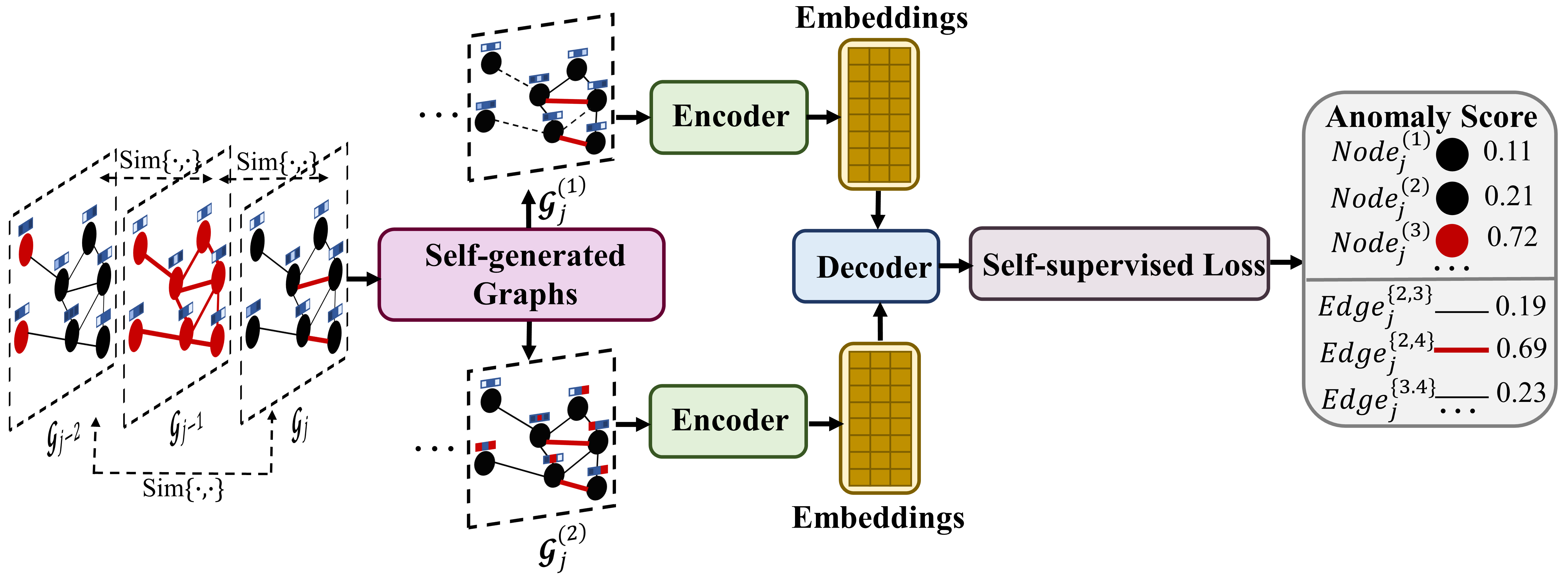}
\\
\caption{Overall framework of Self-supervised methods. The input is a graph $\mathbb{G}$ with three successive observations. In each $\mathcal{G}, m=3$ and the edge features are not shown for simplicity.}
\label{fig:self_supervised}
\end{figure}

\textbf{Overview.} Fig. \ref{fig:self_supervised} and Table. \ref{tab:Selfsupervised_methods} provide a visualization of the overall framework and representative methods in the self-supervised category. Self-supervised learning (SSL), a subset of unsupervised methods \cite{yao2022dota}, has provided novel insights into training without annotated labels \cite{zhang2024self}. Similar to other fields, SSL has demonstrated its effectiveness in graph-based learning methodologies \cite{liu2022graph}. The intuition of SSL is to learn transferable knowledge from massive unlabeled data with well-designed pretext tasks and then generalize the learned knowledge to the downstream tasks \cite{hojjati2022self}. In the SSL-based G-TSAD research, a sub-group of SSL methods, called contrastive learning (CL), has been used so far. CL aims to learn useful representations of data by contrasting positive and negative pairs, i.e., CL is trained to create an embedding space that pulls positive (similar) samples close together and pushes negative (dissimilar) samples far apart. To achieve this, a contrastive loss function is defined to encourage the model to maximize the similarity between positive samples while minimizing the similarity between negative samples. This approach can be formally described as:
\begin{equation}
    \theta^{*}, \phi^{*} = \operatorname*{argmin}_{\theta,\phi} \mathcal{L}_{\text{con}} \Big(D_{\phi}
    \Big(E_{\theta}(\mathbb{G}^{(1)}), E_{\theta}(\mathbb{G}^{(2)})\Big)\Big),
\end{equation}
where $\mathcal{L}_{\text{con}}$ is the contrastive loss, $\mathbb{G}^{(1)}$ and $\mathbb{G}^{(2)}$ are two different sets of graph samples, $D_{\phi}$ is the discriminator that estimates the similarity between these two sets of graph samples. In unsupervised AD, data augmentation is essential to create negative samples/pairs while every two normal samples can potentially form a positive pair. Applying augmentation on graphs is challenging compared to other data types; this will be discussed in Section \ref{sec:discussion}.

AddGraph \cite{zheng2019addgraph} is an unsupervised anomalous edge detection technique that has access to only normal edges during training. It employs CL to train its graphs obtained from social networks, where the positive samples are all the available normal edges while negative samples are generated by applying the Bernoulli distribution to the normal edges. AddGraph employs a temporal graph convolutional network with an attention-based GRU to learn long and short-term temporal patterns of the nodes. The hidden state of the nodes at each time step is used to compute the anomaly scores for all edges. As the generated negative edges may have a possibility of being normal, it is inappropriate to use a strict loss function such as CE; hence, AddGraph proposes a margin-based pairwise loss to distinguishes the positive edges from the generated negative ones.

\begin{table*}%[!ht]
\caption{Summary of Representative Self-supervised Methods.}
\fontsize{7}{8.7}\selectfont
\centering
\begin{threeparttable}
\begin{tabular}{c|c|c|c|c|c|c|c|c}
\hline
\textbf{Method} & \textbf{Year} & \textbf{\makecell{Data \\ Type}} & \textbf{\makecell{Constructed \\ Graph}} & \textbf{\makecell{Learning \\ Task}} & \textbf{Technique} & \textbf{\makecell{Anomaly \\ Type}} & \textbf{\makecell{Evaluation \\ Metric}} & \textbf{\makecell{Targeted \\ Application}} \\ 
 \hline

 AddGraph \cite{zheng2019addgraph}  &  2019 & \makecell{Social \\ Networks} & Dynamic & Edge & \makecell{Temporal GNN and \\ attention-based GRU} & Edge &	AUC & Social Networks \\  \hline
 TADDY \cite{liu2021anomaly} &  2021 & \makecell{Social \\ Networks} &	Dynamic & Edge & \makecell{A graph transformer with an \\ edge-based substructure sampling} &	Edge & AUC & \makecell{Social Networks, \\ Internet Systems} \\  \hline
 CRFAD \cite{purwanto2021dance} &  2021 & Videos & Dynamic & \makecell{Node, \\ Edge} & \makecell{A self-attention module and \\ conditional random fields} & Graph & AUC & 	Surveillance Videos \\  \hline
 CAAD \cite{chang2021contrastive} &  2021 & Videos & Dynamic & \makecell{Node, \\ Edge} & \makecell{A contrastive attention module and \\ a classification module} & Graph & AUC & Real-world Videos \\  \hline 
 EEG-CGS \cite{ho2022self} &  2023 & Signals & Static & -- & \makecell{GNN-based AE and CL modules, \\ with a sub-graph sampling technique} & \makecell{Node, \\ Sub-graph} & Prec, Rec, F1 & Seizure Analysis \\  \hline
 mVSG-VFP \cite{hadi2023vehicle} &  2023 & Signals & Static &  --  & \makecell{GNN-based AE and CL modules \\ with a predefined $\text{Sim} \{\cdot,\cdot\}$}  &	Graph & Prec, Rec, F1 & \makecell{Vehicle Failure \\ Prediction} \\  \hline
 IVAD \cite{doshi2023towards} &  2023 & Videos & Dynamic & \makecell{Node, \\ Edge} & \makecell{A dual-monitoring approach for \\ objects and their interactions} & \makecell{Edge, \\ Graph} & AUC &	Real-world Videos \\  \hline
 STGA \cite{chen2023spatial} &  2023 & Videos & Dynamic & \makecell{Node, \\ Edge} & A multi-head attention GNN & Graph & AUC &	Real-world Videos \\  \hline

\end{tabular}
% \begin{tablenotes}
%   \fontsize{7}{10}\selectfont
%   \item $^{*}$ \footnotesize{App.: Application.}
% \end{tablenotes}
\end{threeparttable}
\label{tab:Selfsupervised_methods}
\end{table*}

Another CL study aiming to detect abnormal edges in social networks and cyber security systems, called TADDY \cite{liu2021anomaly}, also assumes that all normal edges are positive during training, similar to AddGraph. However, TADDY proposes a different negative sampling strategy. Given positive pairs, TADDY randomly generates candidate negative pairs with a number equal to the number of positive pairs. It then checks these candidates to ensure that they are different from the existing normal edge set. Then, TADDY develops a framework with four components: an edge-based substructure sampling to sample the target nodes and the surrounding nodes; a node encoding component to generate the node embeddings in both intra- and inter-variable spaces; a graph transformer to extract the intra- and inter-variable knowledge of edges; and a discriminative anomaly detector based on CE to calculate anomaly scores for all edges. 
 
Although TADDY demonstrates its effectiveness on six real-world graph datasets, while AddGraph uses only two social network datasets, TADDY still remains several issues. First, TADDY implements a random selection technique for generating negative samples, but it does not ensure that they do not belong to the normal set. Hence, its negative sampling approach is less reliable than AddGraph. Second, given the random selection technique of TADDY, the bias can easily occur when the negative sample set is not sufficient to represent the full distribution of negative pairs; hence, additional sampling techniques are required. Lastly, it employs a CE loss to distinguish normal edges from negative edges; however, as denoted in AddGraph, since it is not guaranteed that all generated negative samples are actual non-normal, using a strict loss (e.g., CE) is inappropriate.

The major challenge of both AddGraph and TADDY is that they consider the generated negative samples as actual abnormal samples since there is no access to the actual abnormal data during training, and the generated negative samples are created based on the given normal data. This limits the methods' performance as the generated negative samples might be just a variation of the normal data or a subset of the possible abnormal distribution, resulting in biased representation learning for the actual abnormal data.

Very recent studies \cite{ho2022self,hadi2023vehicle} have addressed these challenges by recognizing that the negative samples generated based on the normal data may differ from the actual abnormal data that will be observed in the test phase. These methods demonstrate that discriminating the positive and negative pairs during training through minimizing the contrastive loss is just for providing a less noisy and more reliable representation space for the \emph{normal} data as all the negative samples are created from normal data -- in other words, the generated negative samples are just a transformation of normal data, not an actual abnormal sample.

EEG-CGS \cite{ho2022self} introduces a sub-graph sampling technique for the epilepsy application of detecting abnormal channels and regions in the brain using multivariate time-series signals. EEG-CGS first represents the signals as static graphs, constructed based on the Euclidean distance or correlation matrices, to capture the relations between sensors. It then proposes a local sub-graph sampling strategy to sample contrastive pairs. For each target node, a positive sub-graph is sampled by closely connected nodes based on controlling the radius of surrounding nodes, while a negative sub-graph is sampled by finding the farthest nodes using the same radius. By sampling positive and negative pairs based on the knowledge of local sub-graphs, EEG-CGS overcomes the issues of the random selection technique seen in TADDY. 

%EEG-CGS then designs two SSL modules, namely reconstruction and contrastive modules\s{, for learning the relationship between a sensor and its surrounding contexts}. The first module is based on GNN-AE, which detects anomalies using feature information. It first anonymizes a node in the positive sub-graphs by replacing its feature vector with an all-zero vector. It then reconstructs the feature vector of the anonymized node from its surrounding nodes in the sub-graph. The reconstruction loss is computed by comparing the reconstructed node with its ground truth. To help learn inter-variable information, a contrastive module is introduced. CL captures the matching of sub-graph pairs, which yields meaningful information for AD. An anomaly score, as a combination of the two components, is assigned to each graph node. 

Similar to EEG-CGS, mVSG-VFP \cite{hadi2023vehicle} incorporates the reconstruction and contrastive modules to detect vehicle failures. However, its module design differs from EEG-CGS. mVSG-VFP first uses a sliding window to generate segments in the time-series vehicle data and then builds mini-batches of segments. Each segment is represented as a static graph, where each sensor is a node, and edges are predefined using Mutual Information between sensor pairs, measuring the statistical dependence between two sensors. Regarding the reconstruction module, mVSG-VFP masks a segment of one sensor at a time and encourages the GNN-AE module to reconstruct it. Regarding the contrastive module, unlike other CL methods that positive and negative samples are created through a contrastive pair sampling strategy, mVSG-VFP leverages a unique characteristic in their vehicle data, i.e., time-series segments recorded during a trip (aka a continuous recording of sensors) are more similar compared to those that are produced during another trip. This acts as a predefined $\text{Sim}\{\cdot,\cdot\}$, providing an additional layer of insight. While not learnable, this $\text{Sim}\{\cdot,\cdot\}$ operates at the segment-label level by utilizing prior knowledge about whether segments belong to the same vehicle trip.

Given this unique property, adjacent segments in the mini-batch are considered as positive samples, and the remaining segments in the same mini-batch are denoted as negative samples. Then, a contrastive loss is applied to pull adjacent segments closer and push them away from the rest of the mini-batch. mVSG-VFP uses the reconstruction error to detect anomalous graph-level segments. Due to the data's property, a detected abnormal segment can imply that its adjacent segments are also anomalies. As such, mVSG-VFP addresses both intra- and inter-variable dependencies, while EEG-CGS only considers the latter. mVSG-VFP also demonstrates superior performance by leveraging $\text{Sim}\{\cdot,\cdot\}$, compared to other graph-based methods like GDN that do not utilize it. The main challenge of mVSG-VFP is that the way of generating contrastive pairs is only applicable to a data compound of operational blocks like vehicle trips, a property that is not available for many applications where using the knowledge of adjacent segments is meaningless.

CL has also been successfully applied for detecting anomalies in videos. 
%
%Unlike the existing studies in the social network domain, such as AddGraph and TADDY, that inefficiently consider the generated negative samples as abnormal data, several existing CL-based video AD methods assume that a few actual abnormal frames are available during training, i.e., the AD process is weakly supervised.
%
%CRFAD \cite{purwanto2021dance} presents a weakly supervised approach for \t{graph-level} AD in \t{surveillance} videos. It starts off with a relation-aware feature extractor to capture multi-scale features in a video using a convolutional neural network. CRFAD then combines a self-attention module and conditional random fields to model the interactions among local and global features across frames as graphs, where nodes are image patches and edges are \t{computed by the pairwise similarity between the corresponding nodes’ features, and are updated through the model's learning process}. Importantly, CRFAD can model object-level graphs while the existing predictive-based methods, discussed in Section \ref{predictive}, can only model frame-level graphs. Such an object-level framework can learn inter-variable interactions among the objects, which are necessary to detect complex movements. Moreover, CRFAD learns the short-term and long-term intra-variable dependencies across frames by employing a modified version of the temporal relational network. A CL scheme is added to broaden the margin between the positive (normal) pairs and the negative (actual abnormal) pairs. 
%
Many existing studies such as CFRAD  \cite{purwanto2021dance} overlook the important matter of developing models with interpretability that can interpret the causes of anomalies. An important application of video AD is to understand and take appropriate actions whenever an anomaly occurs. A response to an abnormal event in a video is usually dependent on its severity, which cannot be easily understandable without having an interpretable model. To address this, IVAD \cite{doshi2023towards} proposes a dual-monitoring approach that monitors individual objects (aka local object monitoring), and their interactions with other objects present in the scene (aka global object monitoring) in surveillance videos.

In the local branch, each object is monitored independently to determine if its action (e.g., body movement) is anomalous. An anomaly score for each frame, based on reconstruction loss, is computed by a GRU-AE. In the global branch, the relationships between objects (e.g., a human, a predicate, and a bike) within a frame are monitored. This creates a scene graph where objects and predicates (i.e., words that relate two objects) are respectively defined as nodes and edges. CL is adopted in this global branch. Positive pairs are selected from the normal training set, while negative pairs are randomly generated and verified to ensure that they do not belong to the positive set. A contrastive loss is employed to minimize the Euclidean distance between positive samples and maximize the distance between positive and negative samples. This loss is also used as an anomaly score for each object in a frame. Finally, an in-depth analysis on both branches is performed to interpret the root cause of anomalies. If the cause of an anomaly is from the global branch, it is interpreted that the edge-level anomaly is caused due to a previously unseen relationship between objects. Otherwise, if the source of an anomaly is from the local branch, it is caused due to an anomalous object action in the graph.

%Overall, IVAD addresses the interpretability challenge seen in previous G-TSAD studies on videos, i.e., IVAD helps to understand the possible change of individual objects and the diverse relationships between different objects. However,\s{ in terms of creating negative samples,} although IVAD can address the issues of AddGraph and TADDY, i.e. it does not use negative samples as actual abnormal samples, IVAD uses a random selection technique. Randomly selecting negative samples is not sufficient to capture the full distribution of the negatives; hence, its sampling technique is less reliable than EEG-CGS and mVSG-VFP, and thus additional sampling strategies are required.

\textbf{Summary.} Existing SSL studies have shown interesting ideas and addressed some problems seen in the other three categories of G-TSAD methods.  For example, SSL methods incorporate a variety of pretext tasks, such as CL and masked reconstruction, to enhance the representation of normal data. These pretext tasks allow SSL to better learn complex patterns, leading to improved AD performance. Additionally, unlike predictive methods, which focus on modeling each frame in video data as a graph node, SSL methods can go beyond this limitation by constructing object-level graphs. This allows SSL methods to capture detailed interactions between objects within the data. However, as SSL is a new topic to G-TSAD, it has not yet confirmed its effectiveness in a variety of applications compared to predictive methods. Moreover, it is suggested that different effective graph augmentations for positive and negative sample generation be employed to improve performance.

In conclusion, while there may be some overlap in the deep feature learning mechanisms and the use of loss functions, such as reconstruction losses, shared across all method categories, the key distinction lies in the specific losses each category minimizes during training. AE-based methods focus purely on reconstruction error of the previous or current input data. GAN-based methods minimize both generator and discriminator losses. Predictive-based methods minimize prediction errors rather than merely reconstructing the input data, and self-supervised methods minimize pretext-based losses.

\section{Datasets and Applications} \label{sec:datasets}
In this section, we present commonly used datasets related to time-series signals, social networks, and videos that have been widely adopted in G-TSAD research. Table \ref{tab:datasets} summarizes these datasets based on their type, characteristics, specific anomalies, and application domains.

%%%%%%%%%%%%%%%%%%%%%%%%%%%%%%%%%%%%%%%%%%%%%%
\begin{table*}%[!ht]
\caption{Summary of commonly used datasets in G-TSAD research.}
\fontsize{7}{8.7}\selectfont
\centering
\begin{threeparttable}
\begin{tabular}{c|c|c|c|c}
\hline
\textbf{Category} & \textbf{Dataset} & \textbf{\#Features} & \textbf{Anomalies} & \textbf{Application} \\ 
 \hline
 \multirow{14}{*}{\textbf{\makecell{Time-\\series \\ Signals}}} & Yahoo S5 \cite{laptev2015s5} & 8 & Spikes, trend changes in traffic and energy  & Real-world scenario monitoring \\ \cline{2-5}
 & NASA \cite{hundman2018detecting} & 25 & Sensor failures, predictive maintenance faults & Machine predictive maintenance \\ \cline{2-5}  
 & SwaT \cite{mathur2016swat} & 51 & Operational faults in industrial control & Water infrastructure monitoring \\ \cline{2-5}  
 & WADI \cite{ahmed2017wadi} & 127 & Leaks, flow anomalies in water distribution & Water distribution monitoring \\ \cline{2-5}  
 & SMAP \cite{su2019robust} & 55 & Unexpected soil moisture readings & Soil moisture monitoring \\ \cline{2-5}
 & SMD \cite{su2019robust} & 38 & Memory leaks, unexpected server downtimes & Internet server monitoring \\ \cline{2-5}
 & UCR  Archive \cite{wu2021current} & varied & Various anomalies across multiple domains & AD in multiple fields \\ \cline{2-5}
 & UCI Repository \cite{bay2000uci} & varied & Broach range of domain-specific anomalies & AD in various domains \\ \cline{2-5}
 & voraus-AD \cite{brockmann2023voraus} & 20 & Pick-and-place operational faults & Robotics and industrial automation \\ \cline{2-5}
 & PhysioNet \cite{goldberger2000physiobank} & 15 & Cariac and physiological signal anomalies & Physiological monitoring \\ \cline{2-5}
 & European Database  \cite{taddei1992european} & 12 & Abnormal ECG waveforms in cardiac conditions & Cardiac abnormality detection \\ \cline{2-5}
 & PTB-XL \cite{wagner2020ptb} & 100 & ECG abnormalities from diverse patient groups & ECG-based heart disease diagnosis \\ \cline{2-5}
 & ICBEB \cite{liu2018open} & varied &  ECG data with abnormal cardiac disorders & Detection of various cardiac disorders \\ \cline{2-5}
 & TUSZ \cite{shah2018temple} & 20 & EEG signals with normal and seizure activity & Seizure detection in epilepsy patients \\ \hline
 \multirow{6}{*}{\textbf{\makecell{Social \\ Networks}}} & UCI Messages \cite{opsahl2009clustering} & 1899 nodes, 20K edges & Unusual message exchanges between users & Messaging graph-based networks   \\ \cline{2-5}
 & Email-DNC \cite{rossi2015network} & 1869 nodes, 24K edges & Irregular emails in political networks & Email network security and AD \\ \cline{2-5}
 & Digg \cite{de2009social} & 30K nodes, 85K edges & Unusual reply behaviors in a social news & User interactions on social platforms \\ \cline{2-5}
 & Bitcoin-Alpha \cite{kumar2016edge} & 3.7K nodes, 24K edges & Trust rating anomalies in Bitcoin trading & Fraud detection in Bitcoin \\ \cline{2-5}
 & Bitcoin-OTC \cite{kumar2018rev2} & 5K nodes, 35K edges & Unusual trust relationships among Bitcoin & Frauds in cryptocurrency networks \\ \cline{2-5}
 & AS-Topology \cite{zhang2005collecting} & 45K nodes, 110K edges & Irregular patterns in Internet systems & AD in Internet infrastructure \\ \hline
 \multirow{4}{*}{\textbf{Videos}} & UCF-Crime \cite{sultani2018real} & 1900 videos & Violent activities in surveillance videos & Surveillance AD in security scenarios  \\ \cline{2-5}
 & Xd-Violence \cite{wu2020not} & 4754 videos & Violent events recorded in diverse scenarios & Violence AD in surveillance footage \\ \cline{2-5}
 & ShanghaiTech \cite{luo2017revisit} & 437 videos & Unusual pedestrian movements & Behavior analysis in outdoor spaces \\ \cline{2-5} 
 & UCSD-Peds \cite{li2013anomaly} & 98 videos & Unusual events on pedestrian walkways &  Pedestrian safety and behavior \\ \hline
\end{tabular}
% \begin{tablenotes}
%   \fontsize{7}{10}\selectfont
%   \item $^{*}$ \footnotesize{App.: Application.}
% \end{tablenotes}
\end{threeparttable}
\label{tab:datasets}
\end{table*}

\subsection{Time-series Signals}
Numerous existing G-TSAD studies have trained models using popular benchmark time-series signal datasets, such as Yahoo S5 \cite{laptev2015s5}, NASA \cite{hundman2018detecting}, SWaT \cite{mathur2016swat}, WADI \cite{ahmed2017wadi}, SMAP \cite{su2019robust}, and SMD \cite{su2019robust}. However, it has been highlighted in recent literature that these benchmarks possess various flaws, including mislabeled ground truth, triviality, unrealistic anomaly density, and run-to-failure bias \cite{wu2021current,wagner2023timesead,koran2024unveiling}. This renders them unsuitable for the robust evaluation and comparison of G-TSAD algorithms. Therefore, improving  benchmarking practices is crucial to ensure a more accurate assessment of G-TSAD methods. \cite{wu2021current} has introduced The UCR Time Series Anomaly Archive, designed for evaluating and benchmarking AD algorithms. \cite{bay2000uci} has also introduced The UCI Machine Learning Repository, which contains datasets that cover a broad range of domains such finance, healthcare, biology, and social sciences. Additionally, \cite{brockmann2023voraus} has recently introduced the voraus-AD dataset for robotic applications. These datasets serve as valuable resources for testing and developing TSAD algorithms. Moreover, recent studies \cite{ho2023multivariate,lai2023open,erkucs2023new} have showcased the superiority of their proposed TSAD methods on several reliable alternatives collected from biomedical domains such as The PhysioNet \cite{goldberger2000physiobank}, The European ST-T database \cite{taddei1992european}, PTB-XL \cite{wagner2020ptb}, ICBEB \cite{liu2018open}, and TUSZ \cite{shah2018temple}. %These datasets consist of anomalies labeled by a consensus of experts. %For example, PTB-XL contains 15-lead electrocardiogram (ECG) recordings from a diverse set of patients. TUSZ is one of the public and largest electroencephalogram seizure datasets, containing normal and seizure brain activities.

\subsection{Social Networks}
As clarified in Section \ref{subsec:visualizations} and Fig. \ref{subfig:fig_user}, social networks, which are represented as dynamic graphs, are also considered as time-series data. This is because there are always fresh persons who enroll in the community network every day/month/year, hence, the relationship between individuals is changing overtime. Detecting anomalies in these dynamic social networks gains significant attention in G-TSAD research. Various datasets on social networks have been used. For instance, UCI Messages \cite{opsahl2009clustering}
is a social network dataset constructed into dynamic graphs, where each node represents a user, and each edge indicates a message exchange between two users. Email-DNC \cite{rossi2015network} is network of emails, where each node indicates a person in the United States Democratic Party, and each edge denotes an email sent from one person to another. Digg \cite{de2009social} is a network dataset collected from a news website, where each node represents a website user, and each edge indicates that one user replies to another user. Bitcoin-Alpha and Bitcoin-OTC \cite{kumar2016edge,kumar2018rev2} are two ``who-trusts-whom" networks of bitcoin users engaged in trading on the Internet platforms, where each node denotes a user, and an edge is present when one user rates another on the platform. AS-Topology \cite{zhang2005collecting} is a network connection dataset collected from autonomous Internet systems, where each node denotes an autonomous system, and each edge indicates a connection between two autonomous systems.

\subsection{Videos}
As illustrated in Section \ref{subsec:visualizations} and Fig. \ref{subfig:fig_video}, video data is considered as the time series as it is a sequence of frames captured over time. The frames are arranged sequentially to create the illusion of motion when played back at a rapid pace. This sequential nature of video data, where each frame depends on the preceding and succeeding frames, makes it inherently a time-series data. Various video datasets have been used in G-TSAD research such as UCF-Crime \cite{sultani2018real},  Xd-Violence \cite{wu2020not}, ShanghaiTech \cite{luo2017revisit}, and UCSD-Peds \cite{li2013anomaly}. Specifically, UCF-Crime is a large-scale dataset that consists of 13 types of anomalies captured by CCTV camera indoors and outdoors during day and night scenarios. Xd-Violence is a large-scale dataset composed of violent videos from diverse scenarios recorded from CCTV cameras. ShanghaiTech is a medium-scale dataset recorded by a CCTV camera in an outdoor location. UCSD-Peds is a small-scale dataset acquired by a camera to record pedestrian walkways.

\section{Evaluation Metrics} \label{sec:evaluation_metric}

Evaluation metrics play a crucial role in assessing the performance of G-TSAD algorithms. This section presents the commonly used evaluation metrics, and discusses several issues of evaluation protocols and potential solutions.

In the domain of time series signals, several metrics such as F1 score (F1), Precision (Pre), and Recall (Rec) are commonly used. However, it is important to highlight that recent critiques \cite{kim2022towards}  have raised concerns about the evaluation protocol employed in many studies \cite{audibert2020usad,xu2018unsupervised,shen2020timeseries}. Specifically, these existing studies compute F1 score after applying a peculiar evaluation protocol, called point adjustment (\texttt{PA}). Generally, \texttt{PA} considers that if at least one point within an anomaly segment is detected as an anomaly, the entire segment is then considered to be correctly detected as an anomaly. Consequently, this practice leads to high F1 scores, yet it results in a high possibility of overestimating the model performance. \cite{kim2022towards} conducted experiments demonstrating that without \texttt{PA}, the performance of existing methods yields no significant improvement over the baseline. Note that the baseline is simply a randomly initialized reconstruction model, such as an untrained autoencoder that consists of a single-layer LSTM. To address this issue, \cite{kim2022towards} proposed a new evaluation protocol, named \texttt{PA\%K}, where \texttt{K} denotes the threshold, to mitigate the overestimation effect of \texttt{PA}. The concept behind \texttt{PA\%K} is to apply \texttt{PA} to a segment only if the ratio of the number of correctly detected anomalies within that segment to its length exceeds the threshold \texttt{K}. Note that \texttt{K} can be selected manually based on prior knowledge.

While \texttt{PA\%K} can address the \texttt{PA} issue, selecting a decision threshold remains non-trivial in TSAD research. There are four scenarios for threshold finding. The first scenario is a common practice where the threshold is selected directly on the \emph{labeled} test set \cite{chen2022deep,ijcai2022p394,zhang2022grelen}, which can lead to biased and unfair evaluations. In the second scenario, the threshold is selected based on a \emph{labeled} validation set that yields the best F1 score. The selected threshold is then applied to the test set \cite{deng2021graph}. However, the first two cases may not align with real-world unsupervised applications where labeled data is unavailable. To tackle this limitation, the third scenario advocates for using an \emph{unlabeled} validation set to select the threshold, which is then applied to the test set \cite{ho2023multivariate}.

The issue of the above threshold-dependent scenarios lies in their evaluation of the model's performance at a single threshold, which may not capture the model's behavior across all possible thresholds. Hence, several methods in the domains of time-series signals, as well as social networks and videos \cite{dai2022graph,lai2023open,zheng2019addgraph,liu2021anomaly,purwanto2021dance,doshi2023towards,cao2022adaptive} have adopted the fourth scenario, employing the \emph{threshold-independent} metrics such as Area Under the Receiver-Operating Characteristic Curve (AUC) \cite{Zhang_2023_CVPR,Yao_2023_CVPR}, and Area Under the Precision-Recall Curve (APR) \cite{davis2006relationship}. This offers a comprehensive assessment of the model's performance by considering multiple thresholds, which allows for a deeper understanding of the trade-offs between Prec and Rec, facilitating better decision-making for a given application. 

Overall, depending on the specific applications at hand and the goals of the evaluation, it is essential to use an appropriate method for selecting threshold as different scenarios may necessitate different approaches, such as supervised, semi-supervised or unsupervised approaches, to ensure accurate and fair evaluations. Moreover, it is advisable to employ a combination of evaluation metrics to comprehensively assess the model's performance.

\section{Challenges and Future Directions} \label{sec:discussion}
Although G-TSAD is a new topic, it has been rapidly growing in popularity, as can be inferred by the number of studies published in the prestigious Artificial Intelligence venues described in Section \ref{sec:methods}. However, detecting anomalies in graphs with evolving graph features and adjacency matrices is challenging, leading to several major concerns in the existing studies. In this section, we analyze these challenges and pinpoint potential directions for future studies.

\textbf{Theoretical Foundation and Explainability}. Despite its great success in various time-series applications, most existing G-TSAD methods are designed and evaluated by empirical experiments without sufficient theoretical foundations to prove their reliability. Plus, they ignore the explainability of learned representations and predicted results -- e.g., what are the important features of nodes and edges, and adjacency matrices in graphs? Is the output (e.g., anomaly score) sufficient to conclude the abnormality of graph objects? These are important concerns for interpreting the model behavior. Therefore, we believe that in addition to the empirical design, providing a solid theoretical foundation and a deep analysis of the learned representations to improve the model's generalization and robustness are crucial. 

\textbf{On Equivalence between Intra-variable and Inter-variable Dependencies}. Capturing both intra- and inter-variable dependencies in time-series data is very important for effective AD. However, many existing methods could not handle this problem. Regarding intra-variable dependency, time-series data often involve long-term intra-variable correlations, yet many existing studies could not tackle this issue. Regarding inter-variable dependency, it is difficult to pre-define the graph features and adjacency matrices for many time-series systems due to the limited prior knowledge of the developer. Recent studies have shown that without pre-defined graphs, graph structure learning techniques \cite{zhu2021survey} can learn the graph information, i.e., the node/edge features and adjacency matrices, which dynamically evolves over time.  This provides a better capturing of the system's underlying mechanism. Hence, a proper integrating of both graph learning techniques and deep sequence models is essential to provide more accurate AD.

\textbf{Graph Augmentation Strategy}. Augmenting graphs is crucial for SSL methods, e.g., generating positive and negative pairs is required in CL-based methods discussed in Section \ref{sec:selfsupervised}. There are many image augmentation methods in the literature \cite{hojjati2022self}. However, due to the dynamic nature of graphs (e.g., intra-variable complexity, non-Euclidean structure), it is not appropriate to directly apply image-based augmentations to the graphs. Most existing SSL studies only consider random sampling techniques, sub-graph sampling, or graph diffusion for augmentation -- this may provide limited diversity and uncertain invariance. Having additional yet effective augmentation techniques such as feature-based (e.g., masking a portion of node/edge features), edge-based (e.g., adding or removing portions of edges), sampling-based (e.g., sampling nodes and their connected edges using importance sampling or knowledge sampling), and adaptive augmentations (e.g., using learned attention schemes) \cite{wu2021self} would be promising directions for G-TSAD studies.  

\textbf{Detection of Multiple Types of Graph Anomalies}. Most of existing methods could only detect either anomalous nodes, edges, sub-graphs, or graphs. However, in many real-world systems, different observations can have different anomaly types; e.g., in some observations, only a single sensor is anomalous, but in other observations, a group of sensors is anomalous. Moreover, although few studies \cite{hadi2023vehicle} use $\text{Sim}\{\cdot,\cdot\}$ to facilitate detecting anomalous nodes, edges, sub-graphs, and graphs, there is no study that targets to detect anomalous $\text{Sim}\{\cdot,\cdot\}$. Since $\text{Sim}\{\cdot,\cdot\}$ presents both short-term and long-term relationships across observations, analyzing $\text{Sim}\{\cdot,\cdot\}$ can be beneficial for understanding the underlying dynamics of time-series, e.g., any anomalous $\text{Sim}\{\cdot,\cdot\}$ can be considered as an unexpected shift of time-series dynamics. Thus, having a model that can detect multiple anomaly types, including $\text{Sim}\{\cdot,\cdot\}$, would be a very interesting research line.

\textbf{Broader Scope of Methodology}. Each category in the G-TSAD studies has its own strengths and drawbacks. AE-based and predictive-based methods are simple to implement as the losses are easy to build, but recovering the input and learning long-term dependencies is a challenging task to realize. Meanwhile, GAN-based methods are difficult to implement due to the adversarial training process that needs a careful selection of hyperparameters, the network architecture, and the optimization algorithm. SSL methods have shown promising results but designing the frameworks, augmentation techniques, and loss functions heavily relies on empirical experiments. These disadvantages may cause poor AD performance. To improve overall AD accuracy, leveraging the complementary strengths of these approaches is important. For example, we can integrate the AE-based and predictive-based models to detect a wider range of anomalies. While the AE-based detects anomalies that occur in the current/past time, the predictive-based model can detect anomalies that occur in the future.

Another example is a hybrid SSL method, e.g., combining the predictive SSL and contrastive SSL modules to achieve better generalization of the learned representations. Moreover, this combined approach can be adapted to different anomaly types; e.g., the contrastive SSL module learns to detect sudden spikes in the time series, while the predictive SSL module can be trained to detect future abnormal trends unseen in the training data. Overall, the hybrid SSL approach can benefit from multiple pretext tasks of the different SSL modules, yet it is important to design an effective joint learning framework to balance each component in the model \cite{liu2022graph}. Given its great potential, hybrid methods would be interesting for further G-TSAD studies. 

\textbf{Datasets and Evaluation Metrics.} As emphasized in Section \ref{sec:datasets}, acknowledging the limitations of current benchmark TSAD datasets is crucial for advancing the field. It is important to invest efforts in curating high-quality benchmark datasets that reflect real-world scenarios and challenges. Moreover, as discussed in Section \ref{sec:evaluation_metric}, the existing challenges with evaluation protocols and metrics underscore the need for a comprehensive evaluation approach. It is essential to employ a combination of evaluation metrics to thoroughly assess model performance for an accurate and fair comparison between methods while improving reproducibility.

\textbf{Limitations of Graph-based Methods to Real-world Scenarios.} As discussed earlier, most existing methods assume that only normal samples are available during training. Graphs, while effective in capturing the underlying normal patterns of data, face difficulties in scenarios where this assumption does not hold. For instance, in open-set scenarios where only a limited amount of labeled anomalous data is available during training \cite{ding2022catching}, graph-based methods may struggle in learning the data patterns due to insufficient abnormal samples. Recent open-set studies \cite{lai2023open} have shown that their non-graph-based method can outperform graph-based methods \cite{zhao2020multivariate}, which assumes that the training data should contain only normal samples.

In contaminated datasets, where anomalies sneak into the normal training data and both the labels and positions of these anomalies are unknown \cite{jiang2022softpatch}, conventional graph-based methods can become ineffective. Recent studies dealing with contaminated data \cite{ho2023multivariate} have compared their method with state-of-the-art graph-based methods \cite{deng2021graph}. \cite{ho2023multivariate} showed that naively applying graph-based methods to contaminated data leads to inaccurate results, as these methods typically assume that the training data contains only normal samples and is free from noise or anomalies. In \cite{ho2023multivariate}, the challenge of data contamination is addressed by incorporating a decontamination phase before the graph learning process. 

Another limitation arises with highly complex and stochastic signals, such as brain or vehicle systems' signals, which are often noisy and unpredictable due to variabilities in brain states, environmental variabilities, drive behavior, and mechanical wear and tear on top of the common sensor and measurement noise. In such scenarios, the recent methods presented in \cite{ho2022self,hadi2023vehicle} suggest constructing static graphs using predefined adjacency matrices. Their experiments show the superior performance of static graphs compared to state-of-the-art dynamic graphs \cite{deng2021graph}. This improved performance could be due to the current limitations of graph learning techniques or the insufficient availability of training data to properly train these more complex methods. Further exploration of these areas remains a promising direction for future research in the field of G-TSAD.

%\s{In such cases, \v{\cite{ho2022self,hadi2023vehicle} construct} static graphs based on predefined adjacency matrices \b{-- for example, based on Euclidean distances of the sensors or correlation matrices of the sensors' features. Their experiments show a better performance of static graphs compared to the SOTA dynamic graphs. Such results can be attributed to the limitations of the existing dynamic graph learning techniques or the lack of enough training data to train the more complex dynamic graph learning techniques. Further investigation of these domains remains as a future work in the field of G-TSAD. } \s{ have \v{proven more effective than the state-of-the-art methods like GDN \cite{deng2021graph}, which dynamically learn the graphs.} This is because dynamic graphs may struggle to effectively model underlying patterns of the data due to the inherent noise and stochasticity in these signals.} }

Finally, the use of $\text{Sim} \{\cdot,\cdot\}$ as an additional layer of supervision remains an underexplored area. To the best of our knowledge, there is only one study, i.e \cite{hadi2023vehicle}, that leverages $\text{Sim} \{\cdot,\cdot\}$ in this way. \cite{hadi2023vehicle} demonstrates that incorporating $\text{Sim} \{\cdot,\cdot\}$ can enhance model performance compared to other graph-based approaches \cite{deng2021graph} that do not make use of this measure. However, in their approach, $\text{Sim} \{\cdot,\cdot\}$ is predefined based on their given vehicle dataset's characteristics; the definition may not be applicable to other real-world datasets. The result of this study confirms the potentials of employing $\text{Sim} \{\cdot,\cdot\}$ in improving AD. Developing methods that can either learn or adapt $\text{Sim} \{\cdot,\cdot\}$ for various applications would offer new research directions for improving graph-based TSAD techniques.

\section{Conclusion} \label{sec:conclusion}
This paper conducts a comprehensive survey of G-TSAD. First, we introduce new concepts in G-TSAD, the major challenges of time-series data, as well as the advantage of representing them as graphs in the context of AD. Then, we present a systematic taxonomy that groups G-TSAD methods, primarily leveraging deep learning architectures, into four categories: AE-based, GAN-based, predictive-based, and self-supervised methods. For each category, we discuss the technical details of the methods, their strengths and weaknesses, as well as comparisons between them. A wide range of practical time-series applications are also introduced. Finally, we point out several limitations from both technical and application perspectives of the current research and suggest promising directions for future works. We hope this survey will serve as a useful reference for follow-up researchers to explore more research in this field.

\bibliographystyle{IEEEtran}
\bibliography{refs}

\ifCLASSOPTIONcompsoc
  \section*{Acknowledgments}
\else
  \section*{Acknowledgment}
\fi

This work is financially supported by the Natural Sciences, Engineering Research Council of Canada (NSERC), Fonds de recherche du Quebec, and the Department of Electrical and Computer Engineering at McGill University. The authors
also wish to acknowledge the partial support of Calcul Quebec and Compute Canada.

\vspace{-1cm}

\ifCLASSOPTIONcaptionsoff
  \newpage
\fi

\end{document}